\documentclass[a4paper,12pt]{artikel3}
\usepackage{mathrsfs}
\usepackage{charter}
\usepackage{natbib}
\usepackage{algorithm,algorithmicx,algpseudocode}
\usepackage{amsmath,amssymb,amsthm}
\usepackage{subfigure,graphicx,array}
\usepackage[urlcolor=blue,linkcolor=blue, citecolor=blue, colorlinks=true]{hyperref}

\def\rH{\mathscr H}

\def\rL{\mathscr L}
\def\rD{\mathscr D}

\def\rN{\mathscr N}
\def\bR{\mathbb R}
\def\bS{\mathbb S}
\def\bD{\mathbb D}

\def\cN{\mathcal N}
\def\cO{\mathcal O}

\def\cP{P}
\def\cC{\mathcal C}
\def\cM{\mathcal M}

\def\cF{\mathcal F}
\def\cG{\mathcal G}
\def\cJ{\mathcal J}
\def\cL{\mathcal L}

\def\kF{\mathfrak F}

\def\kR{\mathfrak R}

\def\f1{\mathbf 1}

\def\fx{\mathbf x}
\def\fy{\mathbf y}
\def\fq{\mathbf q}
\def\fp{\mathbf p}
\def\fs{\mathbf s}
\def\fn{\mathbf n}
\def\fu{\mathbf u}
\def\fv{\mathbf v}

\DeclareMathOperator\diag{diag}
\DeclareMathOperator\sign{sign}
\DeclareMathOperator\Id{I_d}
\DeclareMathOperator\length{Length}
\DeclareMathOperator\Lip{Lip}

\DeclareMathOperator\diver{div}
\DeclareMathOperator\Per{Per}

\providecommand{\keywords}[1]{\textbf{\textit{Keywords:}} #1}
\newtheorem{remark}{Remark}

\author{\textsc{Da Chen$^{1,2}$} and \textsc{Laurent D. Cohen$^{1}$}\\
$^1$University Paris Dauphine, PSL Research University,\\ CEREMADE, CNRS, UMR 7534, 75016 PARIS, France\\
$^2$Centre Hospitalier National d'Ophtalmologie\\ des Quinze-Vingts, Paris, France
 }      
 \title{From Active Contours to Minimal Geodesic Paths: New Solutions to  Active Contours Problems by Eikonal Equations}
 \date{}

\begin{document}
 \maketitle
 \begin{abstract} 
In this chapter, we give an overview of part of our previous work based on the minimal geodesic path framework and the Eikonal partial differential equation (PDE). We show that by designing  adequate Riemannian and Randers  geodesic metrics the minimal paths can be utilized  to search for solutions to  almost all  of the active contour problems and to the Euler-Mumford elastica problem, which allows to blend the advantages from minimal geodesic paths and those original approaches, i.e. the active contours and elastica curves. The proposed  minimal path-based models can be applied to deal with a broad variety of image analysis tasks such as boundary detection, image segmentation and tubular structure extraction. The  numerical implementations for the computation of minimal paths  are known to be quite efficient thanks to the Eikonal solvers such as the Finsler variant of the fast marching method  introduced in~\citep{mirebeau2014efficient}.

 \end{abstract}

\section{Introduction}
The original active contour model proposed by~\citet{kass1988snakes}  is a variational approach for the applications of boundary detection, image segmentation and object tracking.  It is capable of locally minimizing an energy functional for the purpose of detecting an optimal and continuous curve, either open or closed, to depict the image features of interest such as image edges. The energy functional of this model can be established using the image gray levels or gradients.  A broad variety of  region-based active contour approaches have been intensively studied in order to develop the original active contour model~\citep{kass1988snakes}.  They take into account a type of more robust regional homogeneity  penalization way for image segmentation. As a consequence, these region-based variant approaches are able to search for suitable solutions in more complicated situations. 
In the other hand,  the minimal geodesic path model was first introduced by~\citet{cohen1997global} in order to search for  the global minimum of a simplified version of the functional used in the original active contour model~\citep{kass1988snakes}. In essence, this simplified active contour energy functional can be regarded as a weighted curve length and it can be solved through the Eikonal partial differential equation (PDE). The objective of this chapter is to show how the terms that are commonly used in the active contour models can be handled using different kinds of metrics for geodesic paths, in particular Randers metrics.

In its basic formulation, the minimal geodesic path model extracts image features by a globally optimal curve between two prescribed points, usually provided by the user. A minimal path can be tracked through the solution to  the Eikonal PDE. A large number of  well-established Eikonal solvers have been exploited such as the fast marching methods~\citep{sethian1999fast,mirebeau2014anisotropic,mirebeau2014efficient,mirebeau2018fast}. The global optimality and  the efficient solutions lead to a series of successful minimal path-based applications as reviewed in~\citep{peyre2010geodesic}. However, the original  minimal geodesic path model~\citep{cohen1997global} depends only on the first-order derivative term of a curve. Thus  this model cannot impose  curvature to define the regularization term which corresponds to the curve rigidity property. Finally, in the minimal path-based image segmentation scenario, almost all of the existing geodesic metrics~\citep{mille2015combination,benmansour2009fast,appia2011active} are derived from the image boundary features, which may limit their performance on complicated image segmentation tasks. In the remaining of this chapter, we  proposed different geodesic metrics to overcome the aforementioned drawbacks suffered by the existing minimal path models.

In \citep{kimmel2003fast,kimmel2003regularized}, the authors proposed a listing of the active contour energy terms for image segmentation. In our chapter, we propose to revisit Kimmel's chapter~\citep{kimmel2003fast}, named \emph{fast edge integration}, with minimal  path methods to address the active contour problems. 

\subsection{Outline}
In this chapter, we illustrate different types of geodesic metrics  and their respective applications in image analysis based on the minimal path framework and the Eikonal PDE. The structure of this chapter is organized as follows:
\begin{itemize}
	\item In Section~\ref{sec:AC_MP}, we present the preliminary background on the active contour models involving the  approaches respectively driven by edge- and region-based terms. 
	\item In Section~\ref{sec:BasicMP}, we review the existing minimal path models associated with both of the Riemannian  and  Randers metrics, the corresponding Eikonal PDEs for the computation of  the geodesic distance as well as the gradient descent ordinary differential equations (ODEs) for tracking  geodesic curves.
	\item In Section~\ref{sec:AlignmentMinimalPath}, we present two minimal path models for the minimization of the geometric active contour models with alignment terms~\citep{kimmel2003fast,kimmel2003regularized}. We respectively induce a Randers metric and an anisotropic Riemannian metric  from the variants of the asymmetric and symmetric alignment terms. In this case, the active contour problems with different alignment terms are transferred to the corresponding  minimal path problems. 
	\item In Section~\ref{sec:FinslerElastica}, we present a Finsler elastica minimal path model~\citep{chen2017global}  to solve the Mumford-Euler elastica curve problem~\citep{mumford1994elastica}. This is done by using  a strongly anisotropic and asymmetric Randers geodesic metric that is established over an orientation-lifted space. The orientation dimension is used to represent the curve directions. The  introduced Finsler elastica minimal paths can be naturally applied for the applications of retinal imaging and boundary detection.
	\item In Section~\ref{sec:EikonalMinimalPath}, we present a new method that applies the  Randers minimal path model to address the  region-based active contour problems~\citep{chen2016finsler,chen2019eikonal}. The used Randers metric is derived from the region-based homogeneity term as well as the curve length-based regularization term, relying on the solution to divergence equation-constrained minimization problem. In this case, we can  make use of the  Randers Eikonal PDE to solve a broad variety  of  region-based active contour problems. 
\end{itemize}


\section{Active Contour Models}
\label{sec:AC_MP}
The active contour models  originated from the work proposed in~\citep{kass1988snakes} have obtained successful applications in the fields of boundary detection, image segmentation, object tracking and shape modelling. Basically, the goal of an active contour model can be summarized as seeking a regular parametric curve for the  delineation of  the image feature of interest, such as the object boundaries,  by minimizing an energy functional. In general, the minimization of an active contour energy functional can be carried out by the curve evolution scheme. The crucial point for this scheme is the estimation of the gradient descent flow with adequate boundary conditions. Basically, this scheme amounts  to searching for a parametric curve $\cC_{\tau}:(u,\tau)\in[0,1]\times\bR^+_0\mapsto\cC_\tau(u)\in\Omega$, where $u\in[0,1]$ is the curve parameter and $\tau\in\bR^+_0$ represents the (artificial) time and $\Omega\subset\bR^2$ denotes the image domain. The curve $\cC_{\tau}$ is supposed to  converge to the target boundary when $\tau\to\infty$.

\subsection{Edge-based Active Contour Model}
\subsubsection{The Computation of the Image Gradient Features}
Let $I:\Omega\to\bR$ be a gray level  image. The standard Euclidean  gradient of a (smoothed) image $I$ can be defined in conjunction with a Gaussian kernel $G_\sigma$ with variance $\sigma$: 
\begin{equation}
\label{eq:GVF}
\nabla I_\sigma=\left(\partial_{x} G_\sigma\ast I,~\partial_{y} G_\sigma\ast I\right)^T,
\end{equation}
where $\ast$ denotes the convolution operator,  and $\partial_{x} G_\sigma$, $\partial_{y} G_\sigma$ are the first-order partially differentials of the Gaussian kernel $G_\sigma$ respectively along the axes $x$ and $y$. The magnitude $g:\Omega\to\bR^+_0$ of the image gradient $\nabla I_\sigma$ is a scalar-valued function that can be formulated by
\begin{equation}
\label{eq:GrayLevelMagnitude}
g=\|\nabla I_\sigma\|.	
\end{equation}
In the context of boundary detection and image segmentation, the magnitude $g$ often serves as the edge appearance feature map.

For a vector-valued image $\mathbf{I}=(I_1,I_2,I_3)$ in a RGB color space, we first consider a Jacobi matrix $\mathbf{W}_\sigma(\fx)$ of size $2\times3$ to construct the associated image boundary features for a (smoothed) color image $\mathbf{I}$, i.e.
\begin{equation}
\renewcommand{\arraystretch}{1.5}
\label{eq:VectorizedGradient}
\mathbf{W}_\sigma(\fx)=
\begin{pmatrix}
\displaystyle\partial_x{G_\sigma}\ast I_1, &\displaystyle\partial_x{G_\sigma}\ast I_2,&\displaystyle\partial_x{G_\sigma}\ast I_3\\
\displaystyle\partial_y{G_\sigma}\ast I_1, &\displaystyle\partial_y{G_\sigma}\ast I_2,&\displaystyle\partial_y{G_\sigma}\ast I_3
\end{pmatrix}(\fx).
\end{equation}
The edge appearance feature map $g$ for a color image can be estimated using the Frobenius norm of the matrix $\mathbf{W}_\sigma(\cdot)$. In other words, one can build the magnitude $g$ as follows:
\begin{equation}
\label{eq:ColorMag}
g=\sum_{i=1}^3\sqrt{(\partial_xG_\sigma\ast I_i)^2+(\partial_yG_\sigma\ast I_i)^2}.
\end{equation}

\subsubsection{The original active contour model}
\label{subsec:Snakes}
The original edge-based active contour model or the snakes model~\citep{kass1988snakes} aims to  minimize  the following energy functional
\begin{equation}
\label{eq:Snakes}
E_{\rm snake}(\cC)=\int_0^1\left(\alpha_1\|\cC^\prime(u)\|^2+\alpha_2\|\cC^{\prime\prime}(u)\|^2+P(\cC(u))\right)du,
\end{equation}
where $\cC\in H^2([0,1],\Omega)$ is a regular curve with  non-vanishing velocity and  the parameters $\alpha_1$ and $\alpha_2$ are real positive constants  controlling the importance for each component. The functions $\cC^\prime$ and $\cC^{\prime\prime}$ of the forms 
\begin{equation}
 \label{eq:CurveDerivatives}
 \cC^\prime(u):=\frac{d}{du}\cC(u),\quad\cC^{\prime\prime}(u):=\frac{d}{du}\cC^\prime(u),
 \end{equation}
are respectively the first- and second-order derivatives of the curve $\cC$. Note that the squared norm $\|\cC^{\prime\prime}(\cdot)\|^2$ is relevant to the squared curvature of $\cC$.
The function $P:\Omega\to\bR^+_0$, which is referred to as \emph{potential},  is an image data-driven function. It is defined to have low values around the image structures of interest and high values otherwise.  For instances, the potential $P$ can be set as the decreasing function of the image appearance feature map $g$ or can be derived from the image gray levels~\citep{kass1988snakes}.

The curve evolution equation   can be derived from the Euler-Lagrange equation of the energy functional $E_{\rm snake}$, which reads
\begin{equation}
\label{eq:SnakesFlow}
\frac{\partial\cC_\tau}{\partial\tau}=\underbrace{\alpha_1\,\cC_\tau^{\prime\prime}	-\alpha_2\,\cC_\tau^{\prime\prime\prime\prime}}_{\text{Internal Forces}}-\underbrace{\nabla P(\cC_\tau)}_{\text{External Force}},
\end{equation}
where $\cC_\tau^{\prime\prime\prime\prime}(u)$ is the fourth-order derivative of the curve $\cC_\tau$ with respect to $u$. The first two terms in the evolution equation~\eqref{eq:SnakesFlow} obtained from the curve itself is referred to as the internal forces which impose the regularity of the curve,  while  the last term is the external force used to attract the curve to the features (e.g. object edges) of the target. 
 
\subsubsection{Balloon Force}
\label{subsubsec:BalloonForce}
\citet{cohen1991active} introduced an external force, named the balloon force, to relax the demanding initialization requirement of  the original active contours model~\citep{kass1988snakes}. According to~\citep{cohen1991active}, the balloon force $\mathbf{F}_{\rm balloon}$  can be expressed by
\begin{equation}
\label{eq:BalloonForce}
\mathbf{F}_{\rm balloon}=\rN_\cC,	
\end{equation}
where $\rN_\cC$ is the \emph{outward} unit normal to the curve $\cC$. The balloon force  $\mathbf{F}_{\rm balloon}$ is able to drive the curve expanding from the interior region of $\cC$ to the desired features such as the object boundaries. Thus the initial guess should be located inside  the target and  allowed to be far from the boundaries.
 
In conjunction with the balloon force $\mathbf{F}_{\rm balloon}$,  the corresponding curve evolution equation eventually turns out to be
\begin{equation}
\label{eq:BalloonEvolution}	
\frac{\partial\cC_\tau}{\partial\tau}=\alpha_1\, \cC_\tau^{\prime\prime}	-\alpha_2\,\cC_\tau^{\prime\prime\prime\prime}+\alpha_3\,\mathbf{F}_{\rm balloon}(\cC_\tau)-\frac{\nabla P(\cC_\tau)}{\|\nabla P(\cC_\tau)\|},
\end{equation}
where $\alpha_3\in\bR^+$ is a weighting parameter that controls the importance of the balloon force during the curve evolution. 
 
\subsubsection{Geodesic active contour model}
\label{subsubsec:GAC}
 Let $\Lip([0,1],\Omega)$ be the set of all the Lipschitz continuous curves $\cC:[0,1]\to\Omega$. The geodesic active contour model proposed in~\citep{caselles1997geodesic,yezzi1997geometric} takes into account a potential function $P$ to build a  weighted curve length  
 \begin{equation}
 \label{eq:GACLength}
\rL_{\rm geo}(\cC)=\int_0^1	P(\cC(u))\|\cC^\prime(u)\|du.
 \end{equation}
The weighted curve length $\rL_{\rm geo}$ is intrinsic since it is independent of the parameterization of the curve $\cC$. Comparing to the energy functional~\eqref{eq:Snakes} considered in the original snakes model~\citep{kass1988snakes}, the term which relies on the second-order derivative  of the curve has been removed.

An extremum curve that minimizes the weighted curve length $\rL_{\rm geo}$ is a geodesic path. In~\citep{caselles1997geodesic,yezzi1997geometric}, a locally minimizing curve can be generated by the curve evolution scheme. 
Denoting by $\kappa_\tau:[0,1]\to\bR$  the curvature of  $\cC_\tau$, the curve evolution equation obtained through the Euler-Lagrange equation of the weighted curve length $\rL_{\rm geo}$ reads
\begin{equation}
\label{eq:GeodesicFlow}
\frac{\partial\cC_\tau}{\partial\tau}=P(\cC_\tau)\kappa_\tau\cN_\tau-\langle\nabla P(\cC_\tau),\cN_\tau\rangle \cN_\tau,
\end{equation}
where the operator $\langle\cdot,\cdot\rangle$ stands for the Euclidean scalar product, $\cN_\tau$ is the \emph{inward} unit normal to the curve $\cC_{\tau}$, and $\nabla P$ is the Euclidean gradient of the function $P$ with respect to positions.  In the homogeneous regions of an image, the gradients of the potential $P$ tend to vanish, leading to the following  Euclidean curve shortening flow
\begin{equation*}
\frac{\partial\cC_\tau}{\partial\tau}\propto\kappa_\tau\cN_\tau,
\end{equation*}
 where $\propto$ is the positively proportional operator. When the curve $\cC_\tau$ is near an edge at some time $\tau$, the values of $P(\cC_\tau(\cdot))$ are low and the curve evolution will be dominated by the second term of equation~\eqref{eq:GeodesicFlow}. Therefore, it is a good  choice to initialize the curve $\cC_\tau|_{\tau=0}$  surrounding  the target. The drawback of the solution to  the original active contour energy~\eqref{eq:Snakes} is the high possibility of being stuck in a local minimum. The level set method introduced by~\citet{osher1988fronts} is a very popular way for solving the curve evolution problem~\citep{malladi1995shape,caselles1997geodesic, yezzi1997geometric}, but it still can be stopped by unexpected local minima.  This is why the authors in~\citep{cohen1997global} introduced the use of the minimal geodesic  path to find the global minimum of a simplified active contour energy functional, i.e. the weighted curve length~\eqref{eq:GACLength}. The global minimum is attained  by solving a nonlinear partial differential equation, called the Eikonal PDE, which will be briefly introduced in Section~\ref{subsec:CKModel}.

\subsubsection{Geometric active contour models with alignment terms}
The aforementioned geodesic active contour model~\citep{caselles1997geodesic,yezzi1997geometric} as well as the fast implementation approach~\citep{goldenberg2001fast} utilize a direction-independent potential function $P$ to measure the length of a curve. 
\citet{kimmel2003regularized} introduced an anisotropic  variant of the geodesic active contour model based on an image data-driven vector field $V_{\rm align}:\Omega\to\bR^2$, leading to a new weighted curve length formula
\begin{equation}
\label{eq:genralAlignment}
\rL_{\varrho}(\cC)=\int_0^1 \varrho(\langle V_{\rm align}(\cC),\cN \rangle)	\|\cC^\prime\|du,
\end{equation}
where $\varrho$ is a scalar-value function and $\langle V_{\rm align}(\cC),\cN\rangle$ is an alignment measure term. 
For the applications of boundary detection and image segmentation, the vector field $V_{\rm align}$ can be naturally built by the image gradient vector field $\nabla I_\sigma$.

\citet{kimmel2003regularized} suggested two effective  types of  $\varrho$ for the alignment measure term. The first one is to choose  $\varrho(\langle \nabla I_\sigma(\cC), \cN\rangle)=\langle \nabla I_\sigma(\cC), \cN\rangle$ which yields an asymmetric functional 
\begin{equation}
\label{eq:AlignTerm}
\rL_{\rm align}(\cC)=\int_0^1\big\langle \nabla I(\cC), \cN\big\rangle\,\|\cC^\prime\|du.
\end{equation}
The curve evolution equation for the evolving curve $\cC_\tau$ with respect to the functional~\eqref{eq:AlignTerm} can be expressed by
\begin{equation}
\label{eq:AlignMaximizingFlow1}
\frac{\partial \cC_\tau}{\partial \tau}	= \Delta I_\sigma(\cC_\tau)\cN_\tau,
\end{equation}
where $\Delta$ is the  Laplacian operator such that $\Delta I_\sigma=\diver(\nabla I_\sigma)$. The extremum curves $\cC_\tau$ as $\tau\to\infty$ corresponds to a maximizer of the functional  $\rL_{\rm align}$.

The second choice introduced by~\citet{kimmel2003regularized} for the  function $\varrho$ is to set 
\begin{equation*}
\varrho(\langle \nabla I(\cC), \cN\rangle)=|\langle \nabla I(\cC), \cN\rangle|,
\end{equation*}
which yields a robust and symmetric variant  of the asymmetric case~\eqref{eq:AlignTerm} as follows:
\begin{equation}
\label{eq:RobustAlignTerm}
\rL^+_{\rm align}(\cC)=\int_0^1\left|\big\langle \cN,\nabla I_\sigma(\cC) \big\rangle\right|\, \|\cC^\prime\|du.
\end{equation} 
The curve evolution equation in this case has a form  
\begin{equation}
\label{eq:AlignMaximizingFlow2}
\frac{\partial \cC_\tau}{\partial \tau}	=\sign(\langle\nabla I_\sigma,\cN_\tau\rangle)\Delta I_\sigma(\cC_\tau)\cN_\tau.	
\end{equation}

The maximization of  the alignment-based functionals  $\rL_{\rm align}$ and $\rL^+_{\rm align}$ can be respectively solved through the curve evolution equations~\eqref{eq:AlignMaximizingFlow1} and~\eqref{eq:AlignMaximizingFlow2} in conjunction with the level set framework~\citep{osher1988fronts}. An optimal curve generated by means of the evolution equation~\eqref{eq:AlignMaximizingFlow1} (resp. the evolution  equation~\eqref{eq:AlignMaximizingFlow2}) corresponds to the local maximum of the alignment-based functional $\rL_{\rm align}$ (resp. the functional $\rL^+_{\rm align}$). 
In Sections.~\ref{subsec:RandersAlignment} and~\ref{subsec:RiemannianAlignment}, we will introduce a Randers  metric and an anisotropic Riemannian metric respectively for the  global minimization of  the slight variants of the alignment-based functionals $\rL_{\rm align}$ and $\rL^+_{\rm align}$.

\subsection{The Piecewise Smooth Mumford-Shah Model and the Piecewise Constant Reduction Model}
\label{subsec:RegionalActiveContours}
The edge-based active contour approaches mentioned above, which take the image gradients as the edge appearance features for image boundary extraction, are quite efficient and effective models. However, the image gradients are very sensitive to the image  noise and spurious edges, which also likely yield strong edge appearance features. In order to address this issue, the region-based active contour approaches exploit more complicated homogeneity criteria to establish the energy functionals.  Significant examples of these  region-based approaches involve the piecewise smooth  Mumford-Shah model~\citep{mumford1989optimal} as well as the associated variant approaches such as~\citep{zhu1996region,cohen1997avoiding,chan2001active,brox2009local}.

\subsubsection{The Piecewise Smooth Mumford-Shah Model} 
\label{ssubsec:PSMS}
Let $K$ be a closed subset of the image domain $\Omega\subset\bR^2$, which is made up of a finite set of smooth curves. The goal for the original piecewise smooth Mumford-Shah model  is to  seek the set $K$ and a piecewise smooth function $f:\Omega\to\bR$  by minimizing the energy functional
\begin{equation}
\label{eq:MumfordShahSFunctional}
E_{\rm MS}(f,K)=\alpha\int_\Omega (I-f)^2d\fx+\hat\alpha\int_{\Omega\backslash K}\|\nabla f\|^2d\fx+\Per(K),
\end{equation}
where $\alpha$ and $\hat\alpha$ are two positive constants which control the relative importance between different terms, $\Per(K)$ is the perimeter of the set $K$, and  $f$ is referred to as an image data fitting function. The first integration term  defines a similarity measure by estimating the approximation errors between the image gray levels $I$ and the data fitting function $f$. The second integration is a smoothness penalty term that imposes a prior to the function $f$.  

Since the original work in~\citep{mumford1989optimal}, many efforts have been devoted to seeking practical solutions to the minimization of the piecewise smooth Mumford-Shah problem~\eqref{eq:MumfordShahSFunctional}. Among them, the explicit curve evolution scheme is a widely considered tool for minimizing the energy functional~\eqref{eq:MumfordShahSFunctional}, which can be implemented  either by the level set method~\citep{osher1988fronts,chan2001level,vese2002multiphase,tsai2001curve} or by the convex relaxation framework~\citep{chan2006algorithms,bresson2007fast,pock2009algorithm}.

\subsubsection{Piecewise Constant Reduction of the Full Mumford-Shah Model}
\label{ssubsec:PC}
The piecewise constant active contour model~\citep{chan2001active,chan2000active,cohen1997avoiding} is a practical variant  of the full Mumford-Shah model. Instead of  using a piecewise smooth function $f$ to approximate the image data,  the piecewise constant model or the active contours without edges model (ACWE) assumes that the image gray levels within each region can be approximated by the mean intensity value estimated in the corresponding  region. 

Let $\cC$ be a simple and closed curve. We denote by $A_\cC\subset\Omega$ the open and bounded subset enclosed by the curve $\cC$.  In the case of binary-valued image segmentation scenario, the piecewise constant active contour model~\citep{chan2001active,cohen1997avoiding} considers an energy  functional
\begin{equation}
\label{eq:PCModel}
E_{\rm ACWE}(\cC,\mu_{\rm in},\mu_{\rm out})
=\alpha\int_{A_\cC}(I-\mu_{\rm in})^2d\fx+\alpha\int_{\Omega\backslash{A_\cC}} (I-\mu_{\rm out})^2d\fx+\rL_{\rm Euclid}(\cC),
\end{equation} 
where  $\mu_{\rm in}$ and $\mu_{\rm out}$ are respectively the mean gray levels inside and outside the closed curve $\cC$ and the term  $\rL_{\rm Euclid}(\cC)$ stands for the Euclidean curve length of $\cC$ which serves as a regularization term for the active contour model. Note that in the original Chan-Vese piecewise constant model~\citep{chan2001active}, there is a term related to the area of the region $A_\cC$, which has been ignored here. 

\subsubsection{Minimization of the Piecewise Constant Active Contour Model}
\citet{chan2001active} proposed a solution for the  minimization of the active contour energy functional $E_{\rm ACWE}$ based on the  level set method~\citep{osher1988fronts,zhao1996variational}. Basically, the level set formulation relies on  a Lipschitz function $\phi:\Omega\to\bR$ such that the open and bounded subset  $A_\cC$  can be represented by $A_\cC:=\{\fx\in\Omega;\phi(\fx)>0\}$ and its boundary $\cC$ can be identified from the zero level set of $\phi$, i.e. $\cC=\{\fx\in\Omega;\phi(\fx)=0\}$. In numerical implementation, the level set function $\phi$ is often set to be  a signed Euclidean distance function. 

Let  $H$ be the Heaviside function which can be formulated by
\begin{equation*}
H(a)=
\begin{cases}
1,&a\geq0,\\
0,&a<0.
\end{cases}	
\end{equation*}
The characteristic function  of the subset $A_\cC$ can be defined by the Heaviside function of the corresponding level set function. Nevertheless,  one can reformulate the  functional~\eqref{eq:PCModel} using the Heaviside function $H$ and the level set function $\phi$ as follows:
\begin{align}
\label{eq:PCModelLS}
\mathcal{E}_{\rm ACWE}(\phi,\mu_{\rm in},\mu_{\rm out})=&\alpha\int_{\Omega}(I-\mu_{\rm in})^2H(\phi)d\fx\nonumber\\
&+\alpha\int_{\Omega} (I-\mu_{\rm out})^2 (1-H(\phi)) d\fx\nonumber\\
&+\int_\Omega \delta(\phi)\,\|\nabla \phi\| d\fx,	
\end{align}
where $\delta=H^\prime$ is the one-dimensional Dirac measure~\citep{chan2001active}. 

As introduced in~\citep{chan2001active}, the minimization of the functional $\mathcal{E}_{\rm ACWE}$ formulated in Eq.~\eqref{eq:PCModelLS} 
 can be addressed  by iteratively evaluating the following two steps:
 
\noindent(1).~Keeping  the level set function $\phi$ fixed and  minimizing  the functional $E_{\rm ACWE}$  with respect to the scalar values $\mu_{\rm in}$ and $\mu_{\rm out}$ will yield 
\begin{equation}
\label{eq:MinimizationStep1}
\mu_{\rm in}^*=\frac{\int_\Omega I\,H(\phi)d\fx}{\int_\Omega H(\phi) d\fx},\quad\mu_{\rm out}^*=	\frac{\int_\Omega I\,(1-H(\phi))d\fx}{\int_\Omega (1-H(\phi))d\fx}.
\end{equation}
(2). Based on the scalar values $\mu_{\rm in}^*,\,\mu_{\rm out}^*$ derived from Eq.~\eqref{eq:MinimizationStep1}, one solves
\begin{equation}
\label{eq:MinimizationStep2}
\min_\phi\,\mathcal{E}_{\rm ACWE}(\phi,\mu_{\rm in}^*,\mu_{\rm out}^*),
 \end{equation} 
with respect to the level set function $\phi$.

According to~\citep{chan2001active}, finding the numerical solution to the minimization problem~\eqref{eq:MinimizationStep2} requires a regularized function $H_\epsilon$ which approximates the Heaviside function $H$ as $\epsilon\to0$. By computing the Euler-Lagrange equation for the level set function $\phi$, one can obtain the following level set evolution equation 
\begin{equation}
\frac{\partial\phi}{\partial\tau}=\delta_\epsilon(\phi)\left[\alpha(I-\mu^*_{\rm out})^2-\alpha(I-\mu^*_{\rm in})^2-\diver\left(\frac{\nabla\phi}{\|\nabla\phi\|}\right)\right]	
\end{equation}
where $\delta_\epsilon:=H_\epsilon^\prime$ is the regularized version of $\delta$. 

\section{Minimal Paths for Edge-based Active Contours Problems}
\label{sec:BasicMP}
\subsection{Cohen-Kimmel Minimal Path Model}  
\label{subsec:CKModel}
The minimal path model introduced by~\citet{cohen1997global} is a powerful tool  in the fields of image analysis and medical imaging~\citep{cohen2006minimal,peyre2010geodesic}. It is designed to search for the global minimum of the weighted curve length  energy~\citep{caselles1997geodesic,yezzi1997geometric}, measured along  a curve $\cC\in\Lip([0,1],\Omega)$ as follows:
\begin{equation}
\label{eq:isotropicLength}
\rL_{\rm IR}(\cC)	=\int_0^1 (c_1+P(\cC(u))\|\cC^\prime(u)\|\,du
\end{equation}
where $c_1$ is a positive constant used for regularization and the function $P:\Omega\to\bR^+$ is a scalar-valued potential. In the context of boundary detection, the potential $P$  is usually set as a decreasing function of  the image gradient magnitude~\citep{cohen1997global}. 

Given a source point $\fs$ and a target point $\fx$ in the domain $\Omega$, a geodesic path or a minimal path $\cG_{\fs,\fx}\in \Lip([0,1],\Omega)$ with $\cG_{\fs,\fx}(0)=\fs$ and $\cG_{\fs,\fx}(1)=\fx$, is a curve that globally minimizes the length $\rL_{\rm IR}$ among all the paths in the set $\Lip([0,1],\Omega)$. In other words, a geodesic path $\cG_{\fs,\fx}$ can be determined by
\begin{equation}
\label{eq:Geodesic}
\cG_{\fs,\fx}=\underset{\cC\in \Lip([0,1],\Omega)}{\rm{arg\,min}}	\,\rL_{\rm IR}(\cC),~s.t.~
\begin{cases}
\cC(0)=\fs,\\
\cC(1)=\fx.	
\end{cases}
\end{equation}
Tracking such a geodesic path from the source point $\fs$ to a target point $\fx$ relies on a geodesic distance map $\rD_{\fs}:\Omega\to\bR^+_0$. At each point $\fx$, the value of the geodesic distance map $\rD_{\fs}(\fx)$ is equal to  the minimal curve length from $\fs$ to $\fx$
\begin{align}
\label{eq:GeodesicDistance}	
\rD_{\fs}(\fx)&=\inf_{\cC\in\Lip([0,1],\Omega)}\{\rL_{\rm IR}(\cC);\cC(0)=\fs,\,\cC(1)=\fx\}\nonumber\\
&=\rL_{\rm IR}(\cG_{\fs,\fx}).
\end{align} 

It is known that the geodesic distance map $\rD_{\fs}$ is the unique viscosity solution to the Hamiltonian-Jacobi-Bellman equation or the Eikonal equation
\begin{equation}
\label{eq:IsoEikonalPDE}
\begin{cases}
\|\nabla \rD_\fs(\fx)\|=c_1+P(\fx),& \forall \fx\in\Omega\backslash\{\fs\},\\
\rD_\fs(\fs)=0,
\end{cases}	
\end{equation}
where $\nabla\rD_{\fs}$ is the standard Euclidean gradient of the geodesic distance map $\rD_\fs$. 

Solving a gradient descent ODE on the  distance map $\rD_\fs$ can generate a geodesic path, or a minimal path $\hat\cG_{\fx,\fs}$, which connects from  $\fx$ to $\fs$. It  corresponds to a back-propagation processing from the target point $\fx$ to the source point $\fs$. The gradient descent ODE can be expressed by
\begin{equation}
\label{eq:IsoODE}
\frac{d\hat\cG_{\fx,\fs}(u)}{du}=-\frac{\nabla\rD_\fs(\hat\cG_{\fx,\fs}(u))}{\|\nabla\rD_\fs(\hat\cG_{\fx,\fs}(u))\|},
\end{equation}
with boundary condition $\hat\cG_{\fx,\fs}(0)=\fx$. Note that the geodesic curve $\hat\cG_{\fx,\fs}$ is parameterized by its arc-length.  The gradient descent ODE~\eqref{eq:IsoODE} can be numerically solved by Heun’s or Runge-Kutta’s methods.

\begin{figure}[t]
\centering
\includegraphics[height=6cm]{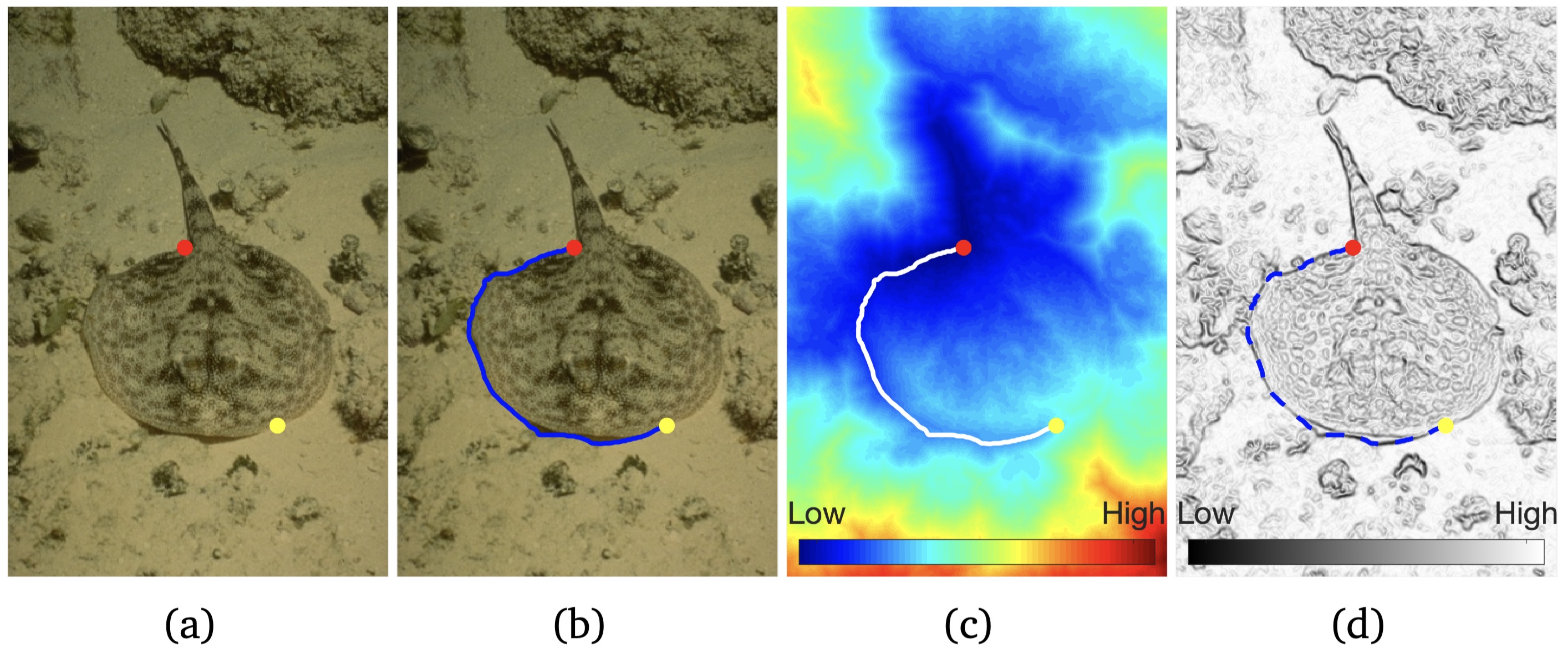}
\caption{An example for the minimal path extraction using the Cohen-Kimmel model~\citep{cohen1997global}. \textbf{a} The source point and the end point which are respectively indicated by red and yellow dots. \textbf{b} The geodesic path indicated by the blue line. This path follows the target boundary between the given points. \textbf{c}  Geodesic distance map. \textbf{d} The potential $P$ with the geodesic path represented by blue dash line.}
\label{fig:IsotropicMinimalPath}		
\end{figure}

\begin{figure}[t]
\centering
\includegraphics[height=5cm]{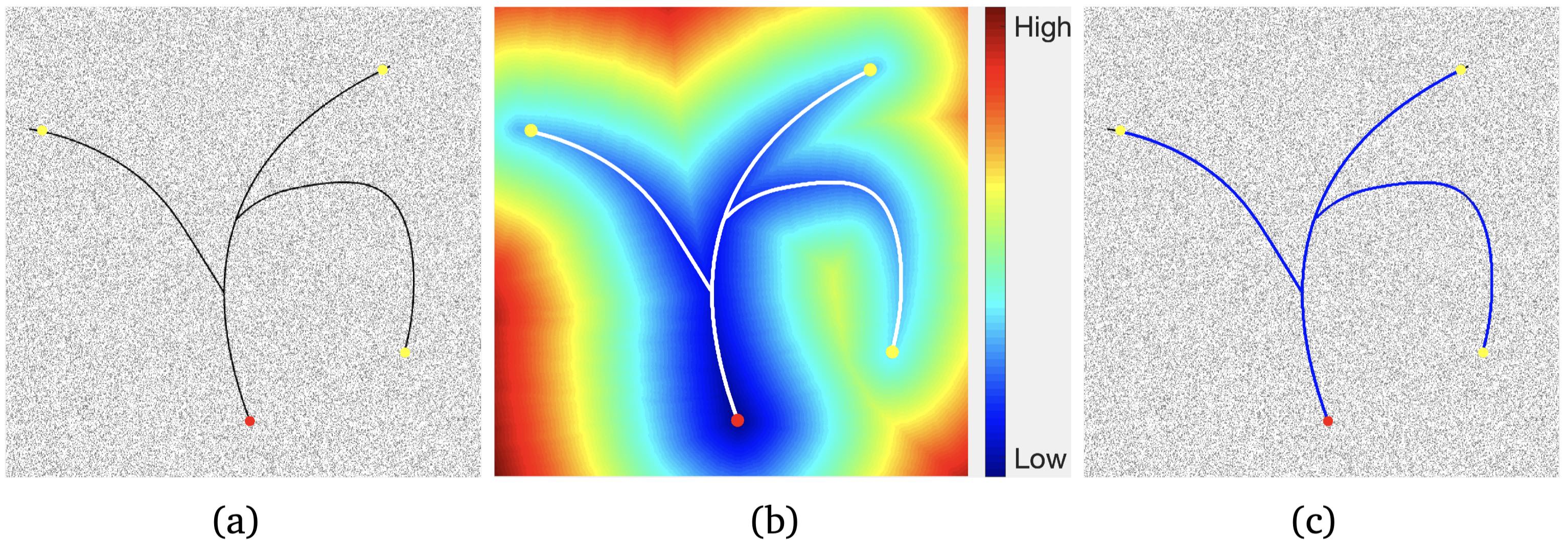}
\caption{An example for the  extraction of a curvilinear tree structure using Minimal paths. \textbf{a} A synthetic image blurred by noise. The red and  yellow dots denote the source point and the end point, respectively. \textbf{b} The geodesic distance map. \textbf{c} The extracted minimal paths which are indicated by blue lines.}
\label{fig:CrackTree}	
\end{figure}

In Fig.~\ref{fig:IsotropicMinimalPath}, we show an example for the computation of a geodesic path between a given pair of prescribed points on an image  from the BSDS500 dataset~\citep{arbelaez2011contour}. In Fig.~\ref{fig:IsotropicMinimalPath}a, the red  and  yellow dots indicate  the source point and the end point, respectively. The corresponding geodesic distance map and the target geodesic path are shown in Figs.~\ref{fig:IsotropicMinimalPath}b and \ref{fig:IsotropicMinimalPath}c, respectively. The potential $P$ is computed by in terms of image gradient features
\begin{equation*}
P(\fx)=\exp(-\beta g(\fx)),	
\end{equation*}
where $\beta$ is a positive constant and $g$ is the magnitude of the color image gradients, see Eq.~\eqref{eq:ColorMag}. We apply the isotropic fast marching method~\citep{sethian1999fast} for the estimation of the  geodesic distance map. 

In Fig.~\ref{fig:CrackTree} we show an example of applying a set of  minimal geodesic paths to extract a curvilinear tree structure. In this example, in order to find the whole tree structure, we give three end points which are located  at the end of each bifurcation. These end points are indicated by yellow dots as illustrated in Fig.~\ref{fig:CrackTree}a. The geodesic distance map and the associated minimal geodesic paths (represented by blue lines) are shown in  Figs.~\ref{fig:CrackTree}b and~\ref{fig:CrackTree}c, repsectively. In this experiment, we build the potential $P$ using the image gray levels as follows:
\begin{equation*}
P(\fx)=c_2+I(\fx),\quad 
\end{equation*}
where the scalar value $c_2\in\bR^+$ is a regularization factor. 

\subsection{Finsler and Randers Minimal Paths}
A Finsler metric is a continuous  map $\kF_{\rm F}:\Omega\times\bR^2\to[0,+\infty]$ over the space $\Omega\times\bR^2$. For each point $\fx\in\Omega$, the Finsler  metric $\kF_{\rm F}(\fx,\vec\fu)$ can be defined through an asymmetric norm, which is a convex and 1-homogeneous function on its second argument. Basically, the asymmetry property of a Finsler metric allows to take advantages of path directions during the computation of the geodesic distance  and of the geodesic paths, as explored in~\citep{melonakos2008finsler,chen2018fast}.  

The weighted  length of a curve $\cC\in\Lip([0,1],\Omega)$ can be measured with respect to a Finsler metric $\kF_{\rm F}$ (see Eq.~\eqref{eq:RandersForm}) by
\begin{equation}
\label{eq:RandersCurveLength}
\rL_{\rm F}(\cC)=\int_0^1\kF_{\rm F}(\cC(u),\cC^\prime(u))	du.
\end{equation}

Similar to the isotropic case defined in Eq.~\eqref{eq:GeodesicDistance}, the geodesic distance map $\rD_{\fs}:\Omega\to\bR^+_0$ associated to the weighted curve length  $\rL_{\rm F}$ can be formulated by 
\begin{equation}
\label{eq:GeodesicDistance}	
\rD_{\fs}(\fx)=\inf_{\cC\in\Lip([0,1],\Omega)}\left\{\rL_{\rm F}(\cC);\cC(0)=\fs,\,\cC(1)=\fx\right\}.
\end{equation} 
In order to estimate the geodesic distance map $\rD_{\fs}$, one can solve  the following Finsler Eikonal equation:
\begin{equation}
\label{eq:RandersEikonalPDE}
\begin{cases}
\displaystyle\sup_{\vec\fv\neq\mathbf{0}}~\frac{\langle\nabla\rD_\fs(\fx),\vec\fv\rangle}{\kF_{\rm F}(\fx,\vec\fv)}=1,&\forall\fx\in\Omega\backslash\{\fs\}\\
\rD_{\fs}(\fs)=0,
\end{cases}	
\end{equation}

The corresponding gradient descent ODE on the geodesic distance map $\rD_\fs$ with respect to the Finsler metric $\kF_{\rm F}$ can be written as 
\begin{equation}
\label{eq:RandersGradientDescentODE}
\begin{cases}
\displaystyle\frac{d\hat\cG_{\fx,\fs}(u)}{du}=\underset{\|\vec\fv\|=1}{\rm{arg\,max}}~\frac{\langle\nabla\rD_{\fs}(\hat\cG_{\fx,\fs}(u)),\vec\fv\rangle}{\kF_{\rm F}(\hat\cG_{\fx,\fs}(u),\vec\fv)},	\\
\hat\cG_{\fx,\fs}(0)=\fx.
\end{cases}
\end{equation}
The target geodesic path $\cG_{\fs,\fx}$ can be computed by re-parameterizing the geodesic path $\hat\cG_{\fx,\fs}$.

In the following, we focus on the Randers metric (denoted by $\kF_{\rm Randers}$) and the Riemannian metric (denoted by $\kR$), which are  two particular cases of Finsler metric. The Eikonal PDEs with respect to these two types of geodesic metrics $\kF_{\rm Randers}$ and $\kR$ are  presented  in Sections~\ref{subsubsec:Randers} and \ref{subsubsec:Riemannian}, respectively.

\subsubsection{Randers Minimal Paths}
\label{subsubsec:Randers}
Let $\bS^+_2$ be the set of all the positive definite symmetric matrices of size $2\times2$. A Randers metric~\citep{randers1941asymmetrical}  is comprised of a positive definite symmetric tensor field $\cM:\Omega\to \bS^+_2$ and a vector field $\omega:\Omega\to\bR^2$
\begin{equation}
\label{eq:RandersForm}	
\kF_{\rm Randers}(\fx,\vec\fu)=\sqrt{\langle\vec\fu,\cM(\fx)\vec\fu\rangle}+\langle\omega(\fx),\vec\fu\rangle,
\end{equation}
where $\langle\cdot,\cdot\rangle$ stands for the standard Euclidean scalar product on $\bR^2$.  A Randers metric $\kF_{\rm Randers}$ of the form~\eqref{eq:RandersForm}  should obey the positive definiteness condition over the domain $\Omega\times\bR^2$. In other words, one has $\kF_{\rm Randers}(\fx,\vec\fu)>0$ for any point $\fx\in\Omega$ and any vector $\vec\fu$,  if and only if 
\begin{equation}
\label{eq:PositiveDefinitess}	
\langle\omega(\fx),\cM^{-1}(\fx)\omega(\fx)\rangle<1.
\end{equation}
For the applications of image analysis, the tensor field $\cM$ and the vector field $\omega$ can be  constructed dependently on the tasks, as formulated in Sections.~\ref{sec:FinslerElastica} and~\ref{sec:EikonalMinimalPath}.

The Eikonal equation~\eqref{eq:RandersEikonalPDE} with respect to a Randers metric $\kF_{\rm Randers}$ is equivalent to the following nonlinear PDE~\citep{mirebeau2017anisotropic,chen2018fast}  
\begin{equation}
\label{eq:RandersEikonalPDE2}
\|\nabla\rD_\fs(\fx)-\omega(\fx)\|_{\cM^{-1}(\fx)}=1,\quad \forall\fx\in\Omega\backslash\{\fs\},	
\end{equation}
with boundary condition $\rD_{\fs}(\fs)=0$.

\subsubsection{Riemannian Minimal Paths}
\label{subsubsec:Riemannian}
The Randers metric $\kF_{\rm Randers}$ of the form~\eqref{eq:RandersForm} gets to be a Riemannian metric $\kR$ of a general form, if the vector field $\omega$ obeys that $\omega\equiv\mathbf 0$, i.e.,
\begin{equation}
\label{eq:RiemannianMetric}
\kR(\fx,\vec\fu)=\sqrt{\langle\vec\fu,\cM(\fx)\vec\fu \rangle}.	
\end{equation}
Then the Eikonal equation  associated to the Riemannian metric $\kR$ defined in Eq.~\eqref{eq:RiemannianMetric} can be formulated by
\begin{equation}
\label{eq:RiemannianEikonal}
\begin{cases}
\|\nabla\rD_\fs(\fx)\|_{\cM^{-1}(\fx)}=1,&\forall\fx\in\Omega\backslash\{\fs\},\\
\rD_\fs(\fs)=0.
\end{cases}	
\end{equation}
In this case, the gradient descent ODE defined in~\eqref{eq:RandersGradientDescentODE} gets to
\begin{equation}
\label{eq:RiemannianODE}
\begin{cases}
\displaystyle\frac{d\hat\cG_{\fx,\fs}(u)}{du}=-\frac{\cM^{-1}(\hat\cG_{\fx,\fs}(u))\nabla\rD_{\fs}(\hat\cG_{\fx,\fs}(u))}{\|\cM^{-1}(\hat\cG_{\fx,\fs}(u))\nabla\rD_{\fs}(\hat\cG_{\fx,\fs}(u))\|},\\
\hat\cG_{\fx,\fs}(0)=\fx.	
\end{cases}	
\end{equation}
This ODE can be solved by Mirebeau's ODE solver~\citep{mirebeau2014anisotropic}. 

\section{Minimal Paths for Alignment Active Contours}
\label{sec:AlignmentMinimalPath}
Inspired by the  geometric active contour models with edge-based  alignment terms~\citep{kimmel2003regularized,kimmel2003fast},  we introduce a Randers metric and an anisotropic Riemannian metric based on the image gradients for the application of boundary detection. These  metrics are induced using the variant forms of the asymmetric and symmetric edge-based alignment terms. 

\subsection{Randers Alignment Minimal Paths}
\label{subsec:RandersAlignment}
We introduce a method, named the Randers alignment minimal path model, for  boundary detection.
The basic idea is to integrate the asymmetric alignment term $\rL_{\rm align}$ formulated in Eq.~\eqref{eq:AlignTerm}, and a Euclidean curve length-based regularization term to establish a new edge-based energy functional 
\begin{align}
\label{eq:VariantAlign}
\cL_{\rm align}(\cC)&=\rL_{\rm Euclid}(\cC)-\beta \rL_{\rm align}(\cC)	\nonumber\\
&=\int_0^1\|\cC^\prime\|du-\beta\int_0^1\big\langle \cN,\nabla I_\sigma(\cC) \big\rangle\,  \|\cC^\prime\|du\nonumber\\
&=\int_0^1\left(1-\beta\big\langle \cN,\nabla I_\sigma(\cC) \big\rangle\right)\|\cC^\prime\|du,
\end{align}
where $\beta\in\bR^+$ is a parameter that controls the relative importance between the regularization term $\rL_{\rm Euclid}$ and the edge-based asymmetric align term $\rL_{\rm align}$. Note that $\nabla I_\sigma$ represents the gradient vector field of the smoothed image $I$, see Eq.~\eqref{eq:GVF}.
  
\begin{figure}[t]
\centering
\includegraphics[height=8.2cm]{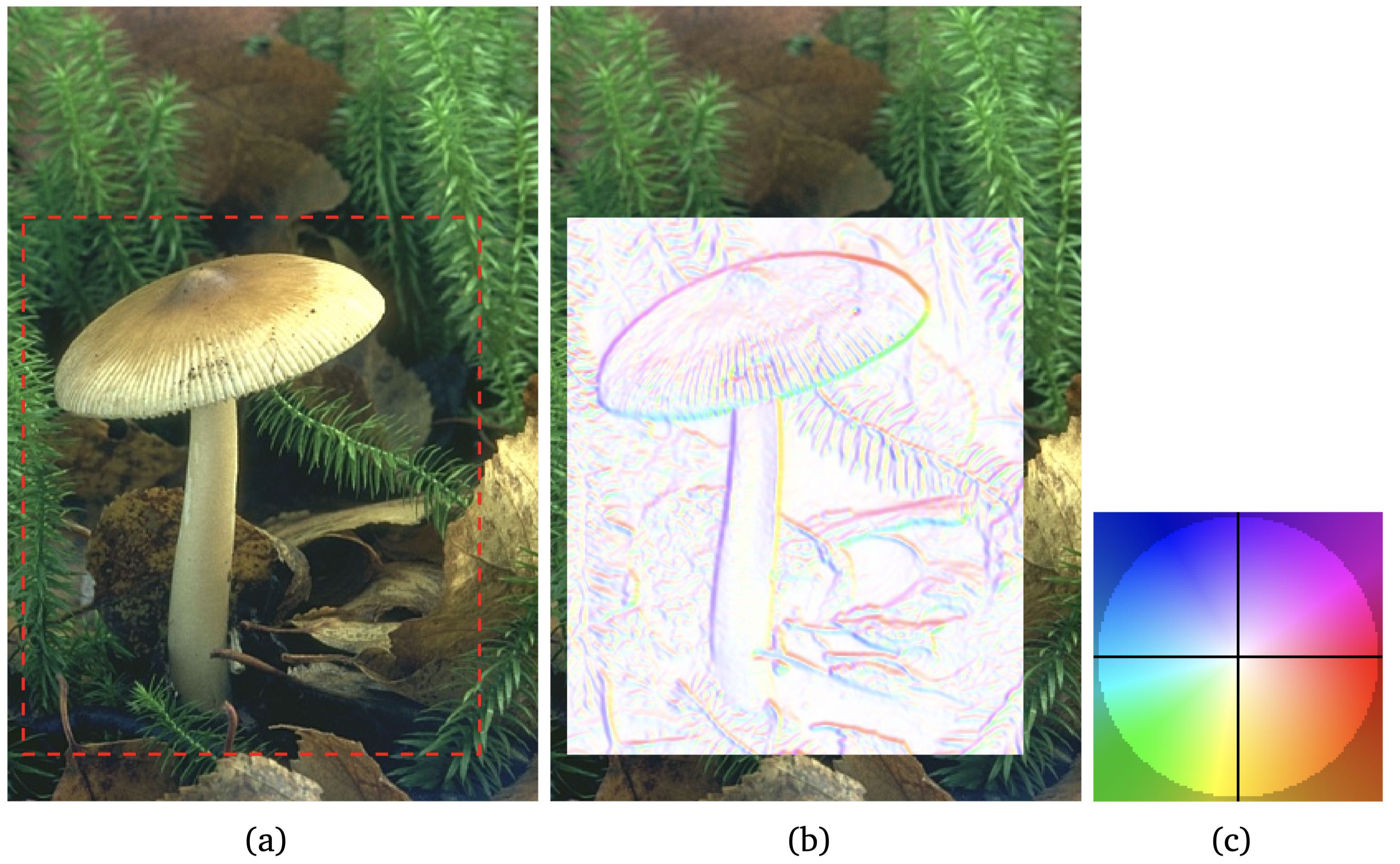}
\caption{Visualization for the vector field $\tilde\xi$. \textbf{a} The original image. \textbf{b} The visualization for the vector field $\tilde\xi$ using colors. We only show the vectors within the rectangle (indicated by red dash line) in (a). \textbf{c} color coding.}
\label{fig:ColorCoding}
\end{figure}

\begin{figure}[t]
\centering
\includegraphics[height=7.5cm]{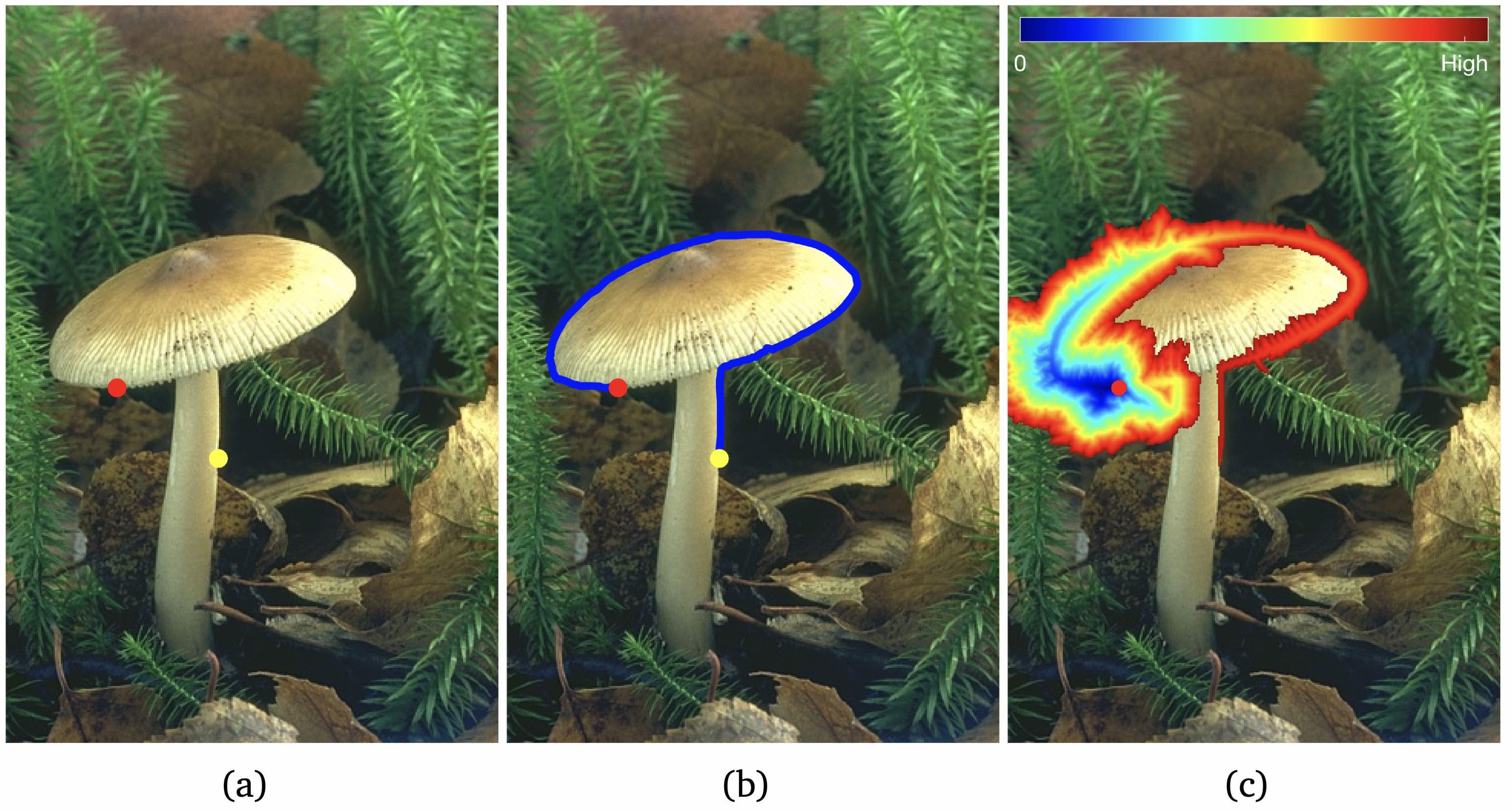}
\caption{An example for boundary detection through an asymmetric Randers alignment minimal path. \textbf{a} The original image and  the seed and end points which are indicated by red and yellow dots, respectively. \textbf{b} The obtained minimal path (indicated by blue line) associated to the Randers metric $\cF_{\rm align}$. \textbf{c} The geodesic distance  map superimposed on the original image.}
\label{fig:RandersAlignment}
\end{figure}

\begin{figure}[t]
\centering
\includegraphics[height=7.5cm]{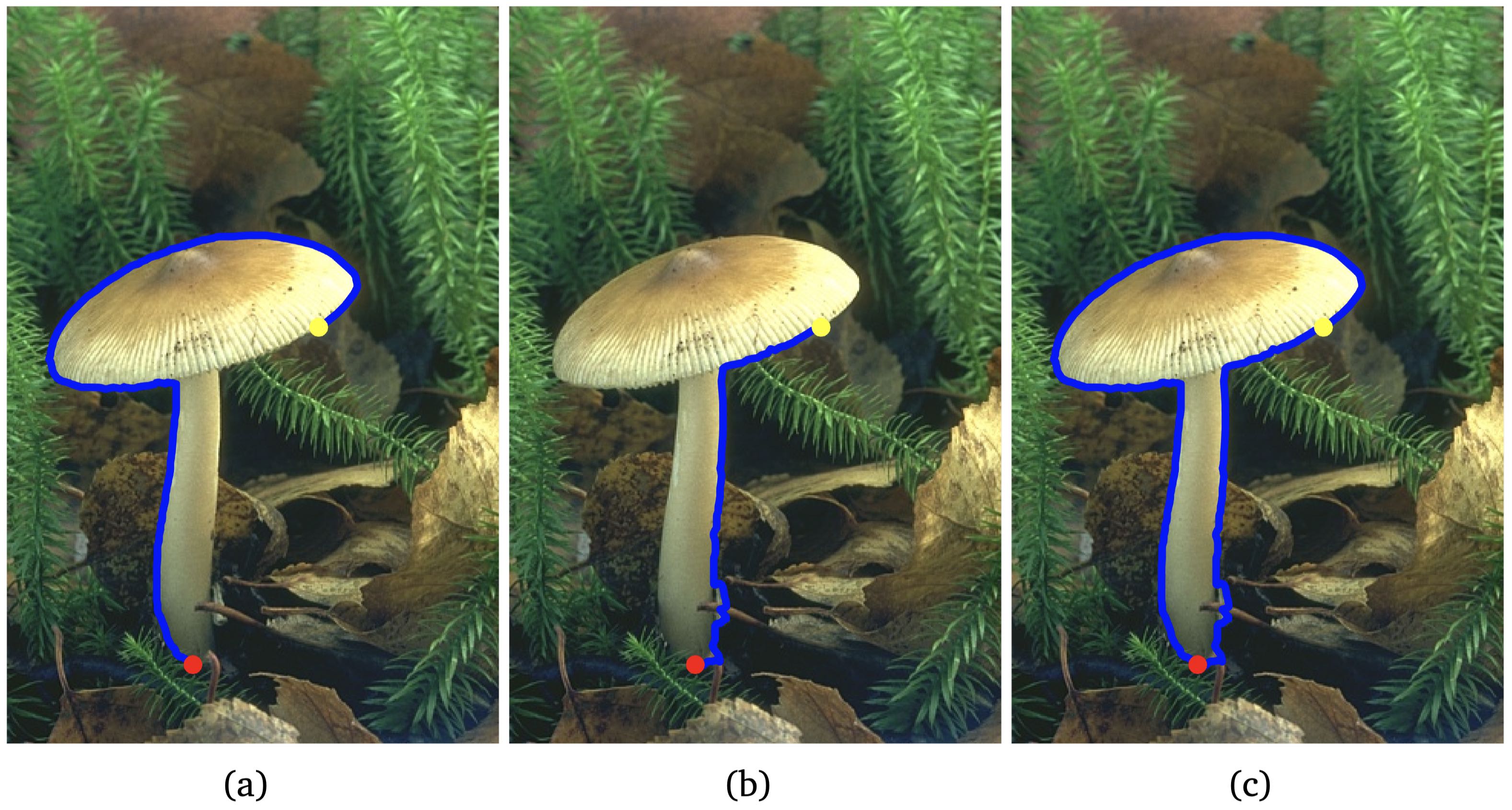}
\caption{The asymmetry property of the Randers alignment minimal paths. \textbf{a} A Randers alignment minimal path generated by respectively taking the red and yellow dots as the source and end points. \textbf{b} Another Randers alignment minimal path generated by exchanging  the source and end points used in (a). \textbf{c} Combining the minimal paths shown in (a) and (b) will generate a closed contour which  passes through the target boundary.}
\label{fig:RandersAlignmentSecond}
\end{figure}

\begin{figure}[t]
\centering
\includegraphics[height=5.5cm]{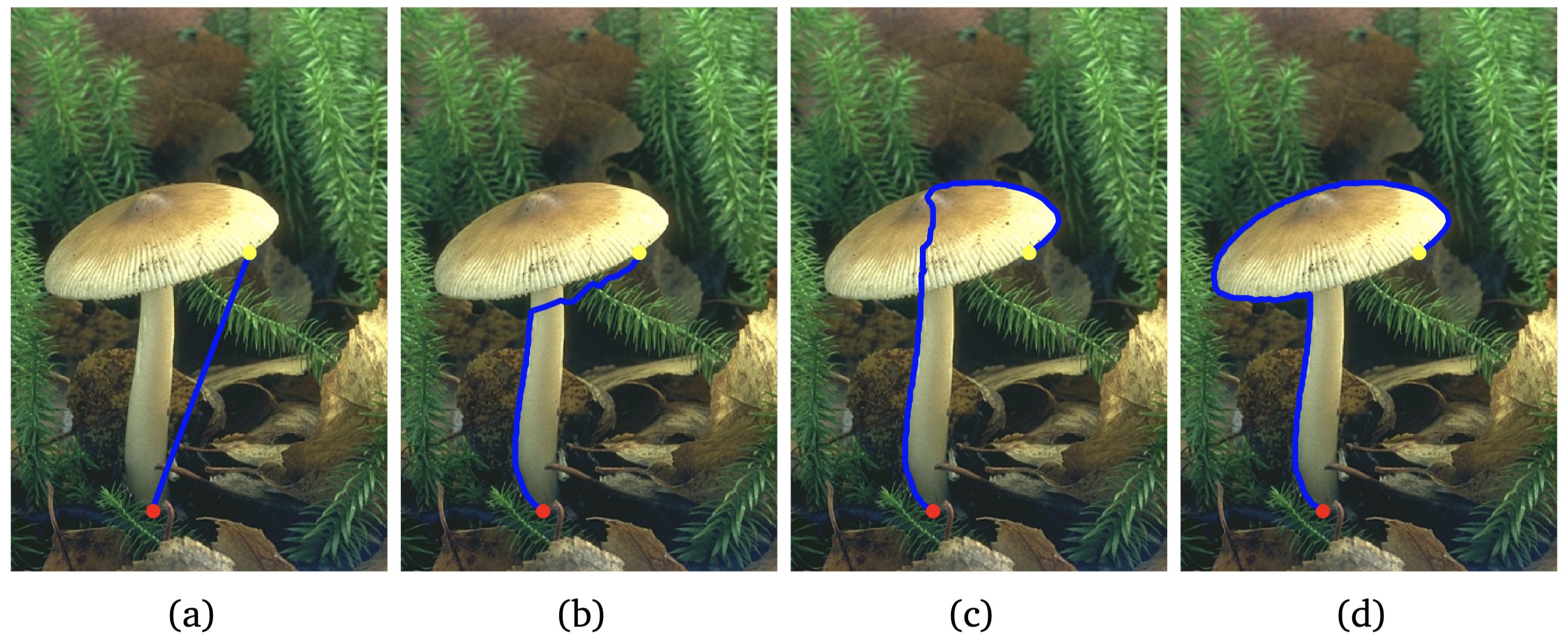}
\caption{The Randers alignment minimal paths associated to different values of $\beta$. \textbf{a} to \textbf{d} show the Randers alignment minimal path extraction results as the values of $\beta$ increase. In particular, we set $\beta=0$ in (a) which yields a straight line segment. In each figure, the red and yellow dots indicate the source point and the end point, respectively. }
\label{fig:RandersAlignmentComparisons}
\end{figure}

The objective in this section is to globally minimize $\cL_{\rm align}$ by searching for a geodesic path between two given points. Let  $\mathbf{M}$ be a rotation matrix with angle $\pi/2$ and let $\xi$  be the rotated image gradient vector field
 \begin{equation}
 \label{eq:ImageTangent}
 \xi(\fx)=\mathbf{M}\,\nabla I_\sigma(\fx).
 \end{equation}
In this case,  a vector $\xi(\fx)$ indicates the direction of a boundary should have at an edge point $\fx$.
 
We reformulate the energy functional $\cL_{\rm align}$ by
\begin{align*}
\cL_{\rm align}(\cC)=&\int_0^1\left(1-\beta\big\langle \mathbf{M}\,\cN,\mathbf{M}\,\nabla I_\sigma(\cC)\big\rangle\right)\, \|\cC^\prime\|du,\\
=& \int_0^1\left(\|\cC^\prime\|-\beta\big\langle \cC^\prime,\xi(\cC) \big\rangle\right)du,\\
=&\int_0^1\kF_{\rm align}(\cC(u),\cC^\prime(u))du,
\end{align*}
where $\kF_{\rm align}$ can be expressed by 
\begin{equation}
\label{eq:AlignmentRander}	
\kF_{\rm align}(\fx,\vec\fu)=\|\vec\fu\|-\langle\vec\fu,\beta \xi(\fx)\rangle.
\end{equation}
Now we can see that $\cL_{\rm align}$  has been formulated as a weighted curve length associated to a Randers metric. The function $\kF_{\rm align}$ is required to obey the positive definiteness condition~\eqref{eq:PositiveDefinitess} in order to globally minimize $\cL_{\rm align}$ through the Eikonal PDE framework. One simple solution is to limit the range of the parameter $\beta$ as follows 
\begin{equation*}
\beta<\left(\inf_{\fx\in\Omega}\|\xi(\fx)\|\right)^{-1}. 	
\end{equation*}
However, a small value of $\beta$ may reduce the importance of the image data, thus increasing the risk of shortcuts. 

As a second choice, we make use of a new vector field $\tilde\xi:\Omega\to\bR^2$ to replace  the term $\beta\xi$ used in the Randers metric $\kF_{\rm align}$. The considered vector field $\tilde\xi$ can be expressed by
\begin{equation}
\label{eq:NonlinearEdge}
\tilde\xi(\fx)=
\begin{cases}
\varphi(\beta \|\xi(\fx)\|)\frac{\xi(\fx)}{\|\xi(\fx)\|},&\text{if~}\|\xi(\fx)\|\neq0,\\	
\mathbf{0},&\text{otherwise}.
\end{cases}
\end{equation}
where $\varphi:\bR^+\to\bR^+$ is a scalar function
\begin{equation}
\label{eq:NonlinearApproximation}
\varphi(a)=1-\exp(-a),\quad\forall a>0. 
\end{equation}

By the vector field $\tilde\xi$, we obtain a new Randers metric $\cF_{\rm align}$ 
\begin{equation}
\label{eq:RobustAlignRanders}
\cF_{\rm align}(\fx,\vec\fu)=	\|\vec\fu\|-\langle\vec\fu,\tilde\xi(\fx)\rangle.
\end{equation}
One can see that the positive definiteness condition characterized by~\eqref{eq:PositiveDefinitess} will always hold for the Randers metric $\cF_{\rm align}$. 

Notice that through the  vector field $\tilde\xi$, we actually minimize  a new weighted curve length  
\begin{align}
\tilde\cL_{\rm align}(\cC)=&\int_0^1\cF_{\rm align}(\cC^\prime(u),\cC(u))\,du\nonumber\\
\label{eq:VariantAlign}
=&\int_0^1\|\cC^\prime\|du-\int_0^1\langle\cC^\prime,\tilde\xi(\cC) \rangle\,du,
\end{align}
where the second term in the second line of Eq.~\eqref{eq:VariantAlign} can be treated as a variant of the asymmetric alignment term $\beta\rL_{\rm align}$.

The asymmetric edge feature at each point $\fx$ still can be characterized  by the vector  $\tilde\xi(\fx)$. Specifically, a large value of $\|\tilde\xi(\fx)\|$  indicates a high possibility that the point $\fx$ belongs to an image edge. Moreover, when $\fx$ is located at an edge,  the direction $\tilde\xi(\fx)/\|\tilde\xi(\fx)\|=\xi(\fx)/\|\xi(\fx)\|$ is proportional to the tangent of the image edge  at point $\fx$.  

In Fig.~\ref{fig:ColorCoding}b, we visualize the vector field $\tilde\xi$ through the color coding scheme~\citep{baker2011database}. In this figure, we only illustrate the vectors $\tilde\xi(\cdot)$ within a rectangle as shown in Fig.~\ref{fig:ColorCoding}a. In Fig.~\ref{fig:ColorCoding}c, we show the color coding, where the direction of a vector is coded by hue while the norm is coded by saturation~\citep{baker2011database}. The color image used in this experiment is obtained  from the BSDS500 dataset~\citep{arbelaez2011contour}.

Notice that when dealing with a vector-valued image $\mathbf{I}:=(I_1,I_2,I_3)$, we build the vector field $\xi$ through the following equation:
\begin{equation*}
\xi=\mathbf{M}\,\left(\sum_{i=1}^3\partial_xG_\sigma\ast I_i,\,\sum_{i=1}^3\partial_yG_\sigma\ast I_i\right)^T.
\end{equation*}

The asymmetry property of the Randers metrics $\kF_{\rm align}$ and $\cF_{\rm align}$ allows to exploit the asymmetric edge features for boundary detection and image segmentation. In Fig.~\ref{fig:RandersAlignment}, we show the boundary detection results using the Randers metric $\cF_{\rm align}$ on a color image. In Fig.~\ref{fig:RandersAlignment}a, the red dot and the green dot respectively denote  the given source and end points. In this experiment, the objective is to extract a geodesic path  to depict the boundary of the object. The geodesic distance map superimposed in the original image is shown in Fig.~\ref{fig:RandersAlignment}c. Note that we adopt the Finsler variant of the fast marching method introduced by~\citet{mirebeau2014efficient} as the Eikonal solver. The computation of the geodesic distance map is terminated once the end point, denoted by the yellow dot, is reached by the geodesic distance front~\citep{deschamps2001fast}.

In Fig.~\ref{fig:RandersAlignmentSecond}, we illustrate the asymmetry property of the Randers alignment minimal paths. In Fig.~\ref{fig:RandersAlignmentSecond}a, the geodesic path is obtained by taking the red dot as the source point and the blue dot as the end point. While in Fig.~\ref{fig:RandersAlignmentSecond}b, we exchange the source and end points used in Fig.~\ref{fig:RandersAlignmentSecond}a.  One can obtain a different minimal path to the one shown in Fig.~\ref{fig:RandersAlignmentSecond}a. Then both minimal paths  can be used to form a closed curve which depicts the complete object boundary, as shown in Fig.~\ref{fig:RandersAlignmentSecond}c.

In Fig.~\ref{fig:RandersAlignmentComparisons}, we show the Randers alignment minimal path  extraction results with respect to different values of the parameter $\beta$. Specifically, we increase the values of $\beta$ from Figs.~\ref{fig:RandersAlignmentComparisons}a to~\ref{fig:RandersAlignmentComparisons}d. In particular, we use $\beta=0$ in Fig.~\ref{fig:RandersAlignmentComparisons}a which generates a straight line segment. 
One can see that a small value of  $\beta$ which  yields a weakly asymmetric Randers metric may generate a geodesic path with  unexpected shortcuts, as shown in Figs.~\ref{fig:RandersAlignmentComparisons}b and~\ref{fig:RandersAlignmentComparisons}c. For a suitable value of $\beta$, the obtained minimal path can successfully depict the object boundary, as shown in Fig.~\ref{fig:RandersAlignmentComparisons}d. 

\begin{figure}[t]
\centering
\includegraphics[height=4.2cm]{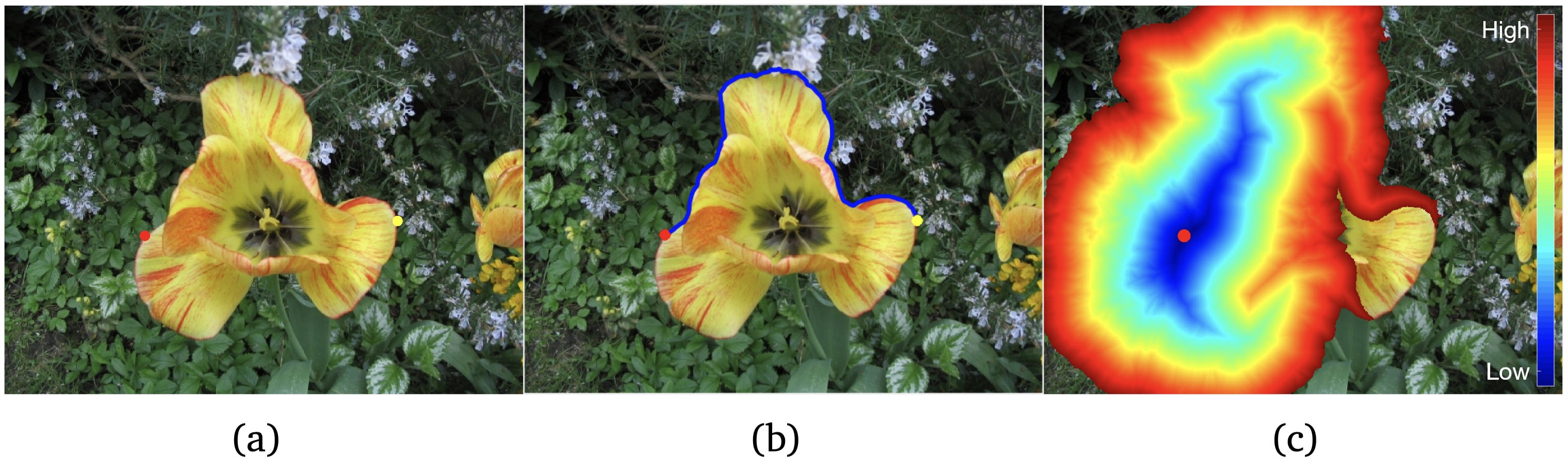}
\caption{Alignment minimal path derived from the Riemannian metric $\mathcal{R}_{\rm align}$. \textbf{a} The original image with the red and green dots respectively indicating the source and end points. \textbf{b} The obtained minimal path between the given points. \textbf{c} The geodesic distance  map superimposed on the original image.}
\label{fig:AlignRiemannian}
\end{figure}

\subsection{Riemannian Alignment Minimal Paths}
\label{subsec:RiemannianAlignment}
In this section, we induce an anisotropic Riemannian metric from a variant form of the symmetric alignment term $\rL_{\rm align}^+$ which is defined in Eq.~\eqref{eq:RobustAlignTerm}. By integrating a  regularization term and the symmetric alignment term, we first introduce an energy functional with a sufficiently small $\beta$
\begin{align}
\cL^+_{\rm align}(\cC)&=\int_0^1\sqrt{1-\beta^2\left|\big\langle \cN,\nabla I_\sigma(\cC) \big\rangle\right|^2}\,\|\cC^\prime\|du\nonumber\\
&=\int_0^1\sqrt{1-\beta^2\left|\big\langle \mathbf{M}\,\cN, \mathbf{M}\,\nabla I_\sigma(\cC) \big\rangle\right|^2}\,\|\cC^\prime\|du\nonumber\\
&=\int_0^1	\sqrt{\|\cC^\prime\|^2-\beta^2\left|\big\langle \cC^\prime,\xi(\cC) \big\rangle\right|^2}\,du\nonumber\\
&=\int_0^1\sqrt{\langle\cC^\prime(u),\cM_{\rm align}(\cC(u))\,\cC^\prime(u)\rangle}\,du.
\end{align}
The tensor field  $\cM_{\rm align}:\Omega\to \bS^+_2$ can be formulated by
\begin{equation}
\label{eq:AlignTensor}
\cM_{\rm align}(\fx)=\Id-\beta^2 \xi(\fx)\xi(\fx)^{T},
\end{equation}
providing that $\beta$ is sufficiently small, where $\xi=\mathbf{M}\,\nabla I_\sigma$ is the rotated image gradient vector field with angle $\pi/2$. In order to ensure the matrix $\cM_{\rm align}(\fx)$ to be positive symmetric definite for any point $\fx$, we replace the term $\beta^2\xi(\cdot)\xi(\cdot)^T$ in Eq.~\eqref{eq:AlignTensor}  by $\tilde{\xi}(\cdot)\tilde{\xi}(\cdot)^T$ defined in Eq.~\eqref{eq:NonlinearEdge}. 

\begin{figure}[t]
\centering
\includegraphics[height=7.3cm]{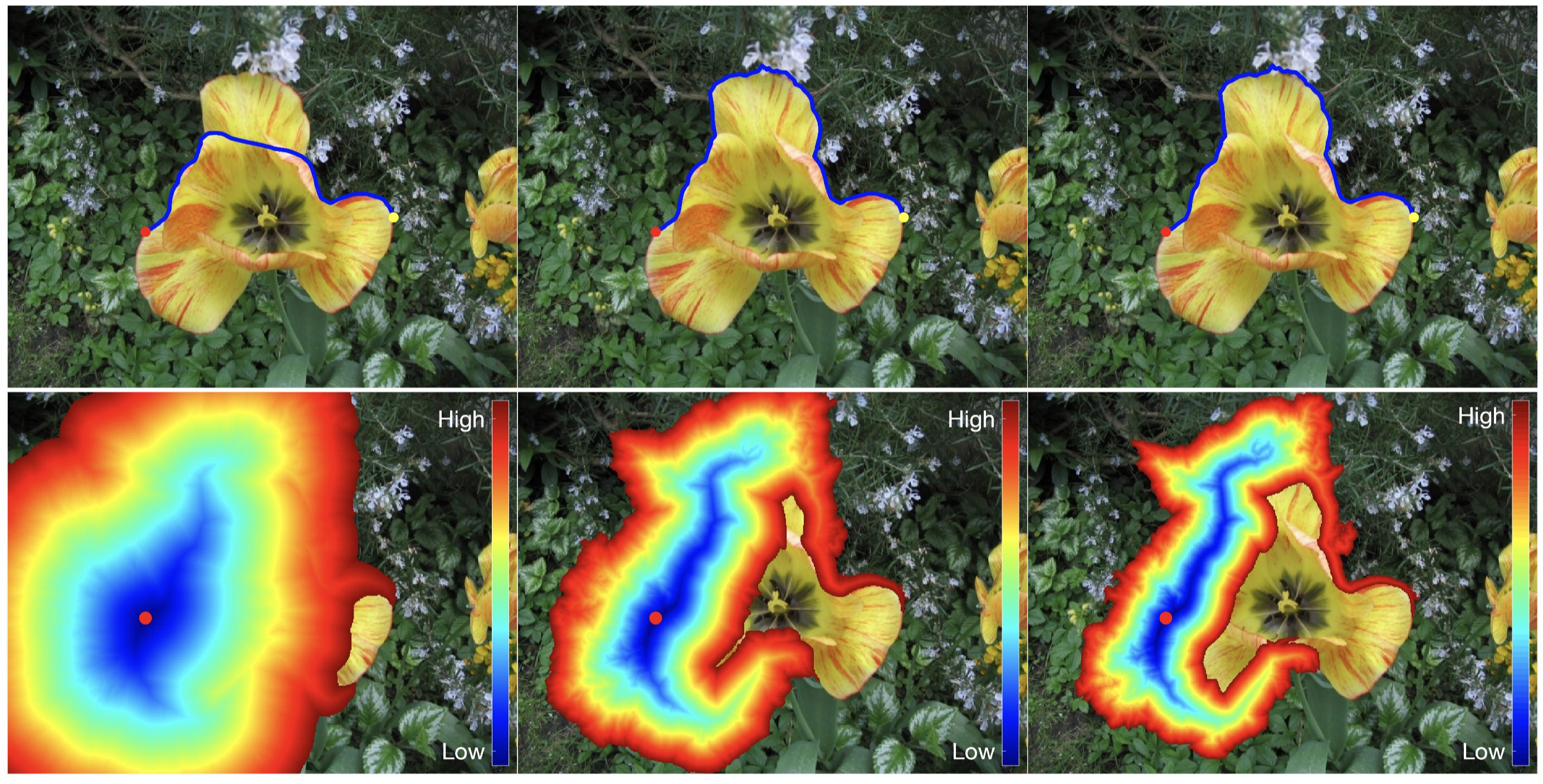}
\caption{The Riemannian alignment minimal paths associated to different values of $\beta$. From columns $1$ to  $3$, we show the results corresponding to  increasing values of $\beta$. The first row shows the minimal paths between two points and the second row illustrates the geodesic distance maps superimposed on the original images. }
\label{fig:AlignRiemannianParameters}
\end{figure}
Hence  we can obtain a new tensor field 
\begin{equation*}
\tilde\cM_{\rm align}(\fx)=\Id-\tilde\xi(\fx)\tilde\xi(\fx)^T,
\end{equation*}
The matrix $\tilde\cM_{\rm align}(\fx)$ will always be  positive definite symmetric for each point $\fx$ since the inequality  $\|\tilde\xi(\fx)\|<1$ always holds. Finally,  we can define a new Riemannian metric
\begin{equation}
\label{eq:RiemannianAlign}
\mathcal{R}_{\rm align}(\fx,\vec\fu)=\|\vec\fu\|_{\tilde\cM_{\rm align}(\fx)}.	
\end{equation}

Following~\citep{sapiro1997color,di1986note}, we take into account the matrix $\mathbf{W}_\sigma(\fx)$ (defined in Eq.~\eqref{eq:VectorizedGradient}) to estimate  the vector $\xi(\fx)$ from a color image $\mathbf{I}$. More precisely, we set 
\begin{equation*}
\xi(\fx)=g(\fx)\upsilon(\fx),\quad\forall\fx\in\Omega
\end{equation*}
where the vector $\upsilon(\fx)$ is the eigenvector  corresponding to the smaller eigenvalue of the matrix $\mathbf{W}_\sigma(\fx)\mathbf{W}_\sigma(\fx)^T$ and $g$ is the Frobenius norm of $\mathbf{W}_\sigma(\fx)$, see Eq.~\eqref{eq:ColorMag}. 

In Fig.~\ref{fig:AlignRiemannian}b, we show  the geodesic path associated to the Riemannian metric $\mathcal{R}_{\rm align}$ on a color image from the Grabcut  dataset~\citep{rother2004grabcut}. The red and yellow dots respectively denote the given source and end  points.  In Fig.~\ref{fig:AlignRiemannian}c, we show the corresponding geodesic distance map superimposed on the original image.

In Fig.~\ref{fig:AlignRiemannianParameters}, we illustrate the Riemannian alignment minimal paths with respect to different values of the parameter $\beta$. In the first row,  from columns $1$ to $3$, the minimal paths indicated by the blue lines are obtained by setting $\beta=4/g_{\rm max},\,8/g_{\rm max}$ and $10/g_{\rm max}$, respectively, where $g_{\rm max}$ is a scalar value defined by 
\begin{equation*}
g_{\rm max}:=\max_{\fx\in\Omega}\|\xi(\fx)\|.	
\end{equation*}
In the first row, we illustrate the respective geodesic distance maps associated to different values of $\beta$. For a small value of $\beta$, we find a shortcut as shown in row $1$ and column $1$.  When increasing the values of $\beta$, one can obtain desired minimal paths which pass through the boundaries of the target, as illustrated in columns $2$ to $3$ of Fig.~\ref{fig:AlignRiemannianParameters}. 

\section{Orientation-lifted Randers Minimal Paths for Euler-Mumford Elastica Problem}
\label{sec:FinslerElastica}

In this section, we denote by $\bS^1=[0,2\pi)$  an orientation space with periodic boundary condition, based on which we can define an orientation-lifted domain $\bD=\Omega\times\bS^1$.  In this case, each point $\hat\fx=(\fx,\theta)\in\bD$ is an ordered pair including a position $\fx$ in the domain $\Omega$ and an orientation  $\theta$ in the orientation space $\bS^1$.

The original Euler-Mumford elastica bending energy functional~\citep{mumford1994elastica,nitzberg19902} assigns to each regular  curve $\cC\in H^2([0,1],\Omega)$ a curvature-dependent length which can be expressed by
\begin{equation}
\label{eq:ElasticaBendingLength}
\rL_{\rm Elastica}(\cC)=\int_0^1(1+\alpha\,\kappa^2(u))\|\cC^\prime(u)\|\,du, 	
\end{equation}
where $\alpha\in\bR^+$ is a scalar-valued parameter which controls the importance of the curvature $\kappa:[0,1]\to\bR$ in  the functional $\rL_{\rm Elastica}$. 

In order to extract the image features such as object boundaries  or tubular structure centerlines, we take into account the following data-driven elastica bending energy~\citep{chen2017global}:
\begin{equation}
\label{eq:DataDrivenBendingLength}
\cL_{\rm Elastica}(\cC)=\int_0^1\Phi(\cC(u),\cC^\prime(u))\big(1+\alpha\,\kappa^2(u)\big)\|\cC^\prime(u)\|\,du, 
\end{equation}
The function $\Phi:\Omega\times\bR^2\to\bR^+$ is a data-driven function relying on both positions and directions. Around the image features of interest and along the proper directions, the orientation-dependent function $\Phi$  is supposed to have small values, and large values, otherwise. The elastica bending energy functional $\rL_{\rm Elastica}$ is a geometric  variant of the original active contour energy~\eqref{eq:Snakes} which also takes the curvature-dependent term for regularization.  The goal of both the Euler-Mumford elastica model~\citep{mumford1994elastica} and the original active contour model~\citep{kass1988snakes} are relevant to one another, i.e. finding an optimal curve of low curvature integration value, which simultaneously tends to  pass through the image features of interest. 

For the sake of simplicity, we will set the data-driven function $\Phi\equiv1$ in Sections~\ref{subsec:FinslerInterpretation} and~\ref{subsec:ElasticaMetric}, and present a Randers minimal path solution to the Euler-Mumford Elastica problem~\eqref{eq:ElasticaBendingLength}. The minimization of the bending energy functional $\cL_{\rm Elastica}$ including the data-driven function will be introduced  in Section~\ref{subsec:DataDrivenElastica}.

\subsection{Euler-Mumford Elastica Problem and its Finsler Metric Interpretation}
\label{subsec:FinslerInterpretation}
The goal  in this section is to search for an optimal  curve which minimizes the elastica bending energy functional 
\begin{equation}
\label{eq:BendingLength}
\rL_{\rm Elastica}(\cC)=\int_0^1(1+\alpha\,\kappa(u)^2)\|\cC^\prime(u)\|\,du,\quad\cC\in H^2([0,1],\Omega),
\end{equation}
by  the Eikonal equation-based minimal path framework. The basic idea is to establish a Finsler geodesic metric to interpret $\rL_{\rm Elastica}$ as a weighted curve length. Since the Eikonal equation is a first-order partial differential equation, we should transform  the squared curvature term to a first-order term via an auxiliary parametric function of orientations. 

We denote by $\varsigma:[0,1]\to\bS^1$ a parametric function over the interval $[0,1]$, where $\varsigma$ is defined being such that
\begin{equation}
\label{eq:TangentRepresentation}
(\cos(\varsigma(u)),\sin(\varsigma(u)))^T=\frac{\cC^\prime(u)}{\|\cC^\prime(u)\|}.
\end{equation}
By a short computation, one can  obtain the following expression 
\begin{equation}
\label{eq:Curva1}
\frac{d}{du}\left(\frac{\cC^\prime(u)}{\|\cC^\prime(u)\|}\right)=\kappa(u)\|\cC^\prime(u)\|\left(\frac{\cC^\prime(u)}{\|\cC^\prime(u)\|}\right)^\perp,
\end{equation}
where $^\perp$ is a perpendicular operator, i.e. $\vec\fu^\perp$ is a vector perpendicular to $\vec\fu$. 

On the other hand, we have the following equations
\begin{align}
\label{eq:Curva2}
&\frac{d}{du}\big(\cos(\varsigma(u)),\sin(\varsigma(u))\big)^T\nonumber\\
=\,&\varsigma^\prime(u)\big(-\sin(\varsigma(u)),\cos(\varsigma(u))\big)^T\nonumber\\
=\,&\varsigma^\prime(u)\left(\frac{\cC^\prime(u)}{\|\cC^\prime(u)\|}\right)^\perp.
\end{align}
In terms of  Eqs.~\eqref{eq:Curva1} and~\eqref{eq:Curva2}, we can express the curvature $\kappa(u)$ by
\begin{equation}
\label{eq:RatioCurvature}
\kappa(u)=\frac{\varsigma^\prime(u)}{\|\cC^\prime(u)\|},\quad \forall u\in[0,1].
\end{equation}
Now we have reformulated  the curvature $\kappa(\cdot)$ as a ratio of two first-order terms $\varsigma^\prime(\cdot)$ and $\|\cC^\prime(\cdot)\|$.  By incorporating the formula in Eq.~\eqref{eq:RatioCurvature} to the elastica bending length $\rL_{\rm Elastica}$, one obtains that
\begin{equation}
\label{eq:BendingLength2}
\rL_{\rm Elastica}(\cC)=\int_0^1 \left(\|\cC^\prime(u)\|+\frac{\alpha|\varsigma^\prime(u)|^2}{\|\cC^\prime(u)\|}\right)\,du,
\end{equation}
following the constraint in Eq.~\eqref{eq:TangentRepresentation}.

Let $\Gamma=(\cC,\varsigma)\in\Lip([0,1],\bD)$ be a canonical orientation-lifted curve and one has $\Gamma^\prime(u)=(\cC^\prime(u),\varsigma^\prime(u)),\,\forall u\in[0,1]$. Now the elastica bending energy functional $\rL_{\rm elastica}$ of the form~\eqref{eq:BendingLength2} can be rewritten by 
\begin{equation}
\label{eq:BendingLength_Metric}
\rL_{\rm Eastica}(\cC)=\int_0^1\kF_\infty(\Gamma(u),\Gamma^\prime(u))\,du,
\end{equation}
where $\kF_\infty:\bD\times\bR^3\to[0,+\infty]$ is an orientation-lifted  Finsler metric. It is defined for any orientation-lifted point $\hat\fx=(\fx,\theta)\in\bD$ and any vector $\tilde{\fu}=(\vec\fu,\nu)\in\bR^3$ by
\begin{equation}
\label{eq:OrienLiftedInfFinsler}
\kF_\infty(\hat\fx,\tilde\fu)=
\begin{cases}
\|\vec\fu\|+\frac{\alpha|\nu|^2}{\|\vec\fu\|},\quad &\text{if~}\vec\fu\propto \vec\fp_\theta\\
+\infty	,\quad&\text{otherwise},
\end{cases}
\end{equation}
where $\vec\fp_\theta=(\cos\theta,\sin\theta)^T$ is a unit vector associated to an orientation $\theta\in\bS^1$ and where $\vec\fu\propto \vec\fp_\theta$ means that the vector $\vec\fu$ is positively proportional to the vector  $\vec\fp_\theta$, i.e., $\vec\fu=\|\vec\fu\|\vec\fp_\theta$.

\subsection{Finsler Elastica Geodesic Path for Approximating the Elastica Curve}
\label{subsec:ElasticaMetric}
The orientation-lifted Finsler metric $\kF_\infty$ defined in Eq.~\eqref{eq:OrienLiftedInfFinsler} involves the infinite value $+\infty$, which will introduce difficulties to the  numerical computation of the geodesic distance maps and the associated  minimal geodesic paths. In order to address this problem, here we consider  a relaxed  orientation-lifted Randers metric $\kF_\lambda$ with a scalar factor $\lambda\gg1$, which can be expressed by
\begin{equation}
\label{eq:FinslerElasticaMetric}
\kF_\lambda(\hat\fx,\tilde\fu)=\sqrt{\lambda^2\|\vec\fu\|^2+2\alpha\lambda|\nu|^2}-(\lambda-1)\langle \vec\fu,\vec\fp_\theta\rangle,
\end{equation} 
where $\alpha\in \bR^+$ is a positive weighing parameter  as  used in Eq.~\eqref{eq:OrienLiftedInfFinsler}. In the following, we refer to the Randers metric $\kF_\lambda$ as the \emph{Finsler elastica metric}.

The metric $\kF_\lambda$ can also be reformulated in the form~\eqref{eq:RandersForm} through a tensor field $\cM_\lambda:\bD\to \bS^+_3$ and a vector field $\omega_\lambda:\bD\to\bR^3$, which can be respectively formulated for any point $\forall\hat\fx=(\fx,\theta)$ as
\begin{equation}
\cM_\lambda(\hat\fx)=\diag(\lambda^2,\lambda^2,2\alpha\lambda),\quad
\omega_\lambda(\hat\fx)=(\lambda-1)(\cos\theta,\sin\theta,0)^T.
\end{equation}
One can claim that the positive definiteness condition formulated in Eq.~\eqref{eq:PositiveDefinitess} can be satisfied:
\begin{equation*}
\langle\omega_\lambda(\hat\fx),\cM^{-1}_\lambda(\hat\fx) \omega_\lambda(\hat\fx)\rangle=(1-\lambda^{-1})^2<1.	
\end{equation*}

Now we give the analysis for the relation between the orientation-lifted Finsler metric $\kF_\infty$ and the Finsler elastica metric $\kF_\lambda$. As $\lambda\to\infty$, we can express the metric $\kF_\lambda$ as follows: 
\begin{align}
\kF_\lambda(\hat\fx,\tilde\fu)&=\sqrt{\lambda^2\|\vec\fu\|^2+2\alpha\lambda|\nu|^2}-(\lambda-1)\langle \vec\fu,\vec\fp_\theta\rangle\nonumber\\	
&=\lambda\|\vec\fu\|\left(\sqrt{1+\frac{\alpha|\nu|^2}{\lambda\|\vec\fu\|}}+\cO(\lambda^{-2})\right)-(\lambda-1)\langle\vec\fu,\vec\fp_\theta\rangle\nonumber\\
&=\|\vec\fu\|+\frac{\alpha|\nu|^2}{\|\vec\fu\|}+(\lambda-1)(\|\vec\fu\|-\langle \vec\fu,\vec\fp_\theta\rangle)+\mathcal O(\lambda^{-1})\nonumber
\end{align}

The value of the term $(\lambda-1)(\|\vec\fu\|-\langle \vec\fu,\vec\fp_\theta\rangle)$ will get to $0$ when the  directions  $\vec\fu$ and  $\vec\fp_\theta$ are  positively proportional to each other. Thus, one can see that $\kF_\lambda \to \kF_\infty$ pointwisely, as $\lambda \to\infty$.

\begin{figure}[t]
\centering
\includegraphics[height=5cm]{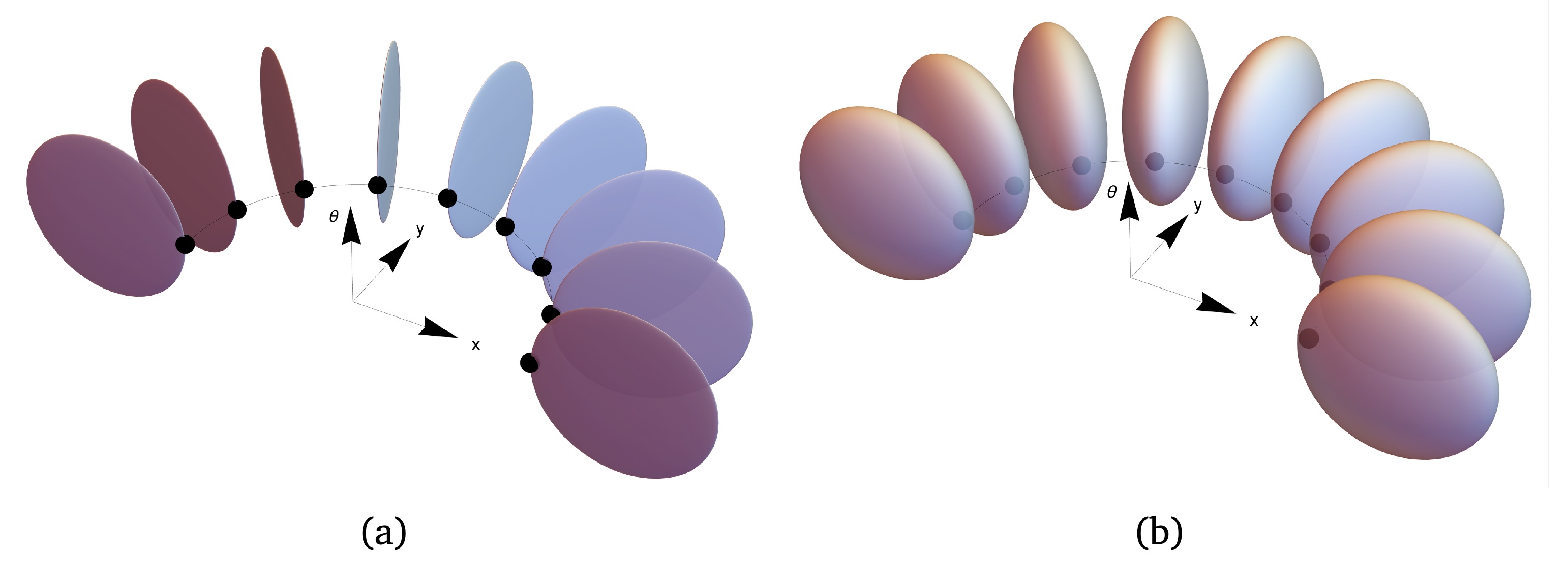}
\caption{Control sets derived from the metrics $\kF_\infty$ and $\kF_\lambda$.  \textbf{a} Control sets of the metrics $\kF_\infty$ associated to different orientations $\theta$ are flat 2D disks embedded in 3D space, which aligned with the direction $\vec\fp_\theta$. \textbf{b} Control sets of the Finsler elastica metrics $\kF_\lambda$ are ellipsoids. We set $\alpha=1$ for the construction of the metrics $\kF_\infty$ and $\kF_\lambda$. The figures are obtained from~\citep{chen2017global}}
\label{fig:Tissot}
\end{figure}

\begin{figure}[t]
\centering
\includegraphics[height=7cm]{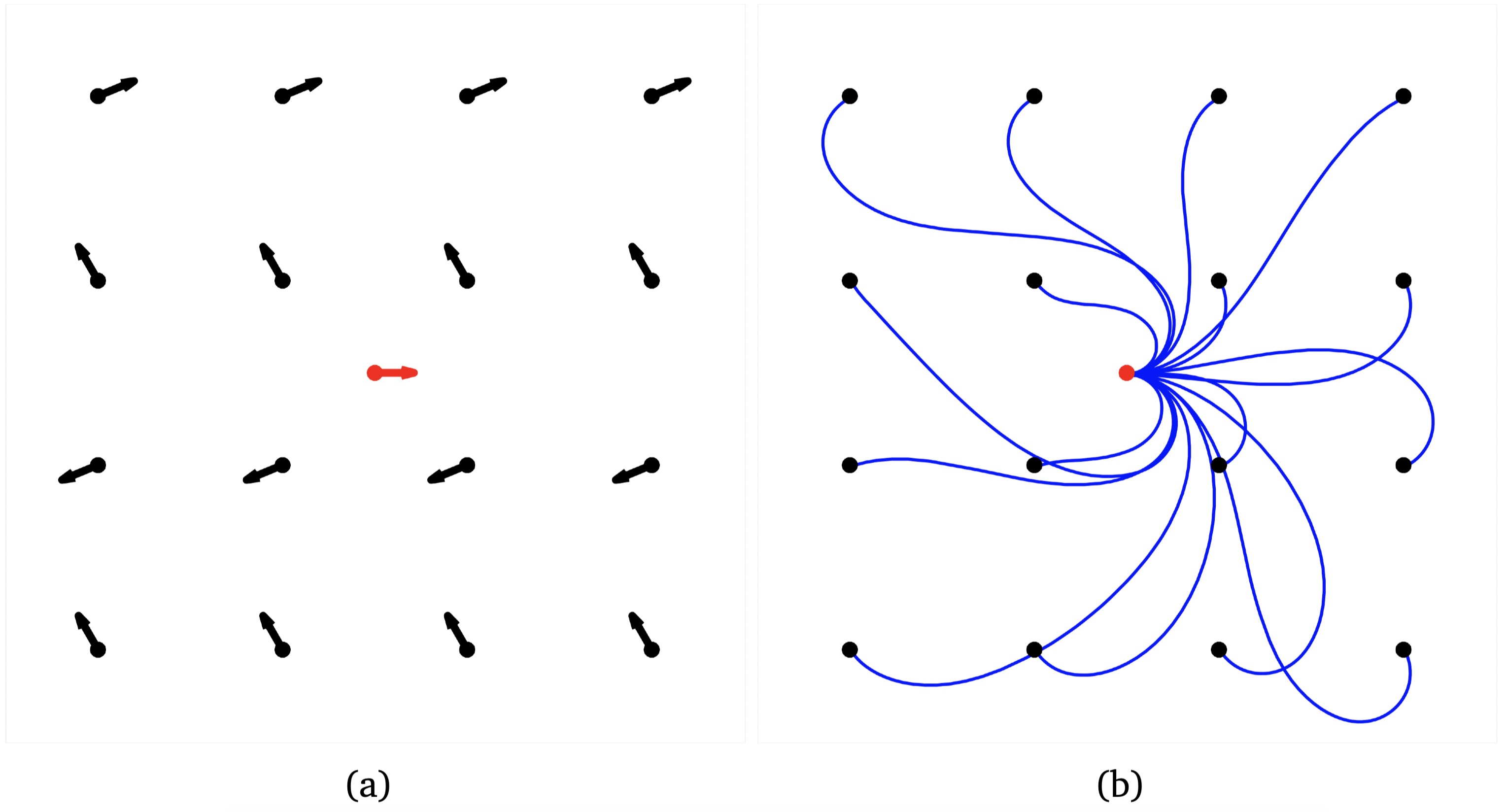}
\caption{Finsler elastica curves associated with the metric $\kF_\lambda$. \textbf{a} The  source position and the end positions are denoted by the red and black dots respectively. Each of the positions is assigned a direction indicated by a arrow. \textbf{b} Minimal Paths (blue lines) associated to the initializations shown in \textbf{a}.}
\label{fig:UniformElasticas}
\end{figure}

The control set or the unit ball is a basic tool for  visualizing  the geometry distortion of  a metric. For the metrics $\kF_\infty$ and $\kF_\lambda$, the sets of the unit balls can be respectively defined as
\begin{equation}
\label{eq:Ball_infty}
B_\infty(\hat\fx)= \{\tilde\fu= (\vec\fu, \nu)\in \bR^3;\,  \kF_\infty(\hat\fx,\tilde\fu) \leq 1\},
\end{equation}
and
\begin{equation}
\label{eq:Ball_lambda}
B_\lambda(\hat\fx)=\{ \tilde\fu = (\vec\fu, \nu)\in \bR^3;\,  \kF_\lambda (\hat\fx,\tilde\fu) \leq 1\}.
\end{equation}
In order to characterize the sets $B_\infty$ and $B_\lambda$, we first define two scalar values $\mu_1$ and $\mu_2$ as follows~\citep{chen2017global}
\begin{equation*}
\mu_1= \langle\vec\fu,\vec\fp_\theta\rangle,\quad
\mu_2 = \langle\vec\fu, \vec\fp_\theta^\perp\rangle.
\end{equation*}

Hence any vector  $\tilde\fu=(\vec\fu,\nu)\in B_\infty(\hat\fx)$ can be characterized by the following inequalities
\begin{equation}
\label{eq:Character}
\mu_1>0, \quad \mu_2 = 0,	\quad \text{and}\quad \mu_1+\alpha\,\frac{|\nu|^{2}}{\mu_1} \leq 1.
\end{equation}
Taking into account the Eq.~\eqref{eq:Character}, one obtains
\begin{equation}
\label{eq:Charac_infty}   
\left(\mu_1 - \frac{1}{2}\right)^2 + \alpha\,|\nu|^{2} \leq \frac{1}{4}.
\end{equation}
Thus the ball $B_\infty(\hat\fx)$ can be characterized by a flat 2D ellipse which is embedded in the 3D tangent space with the origin being on its boundary. In particular, if one sets $\alpha =1$, the unit ball $B_\infty(\hat\fx)$ gets to be a flat 2D disk of radius $1/2$  as shown in Fig.~\ref{fig:Tissot}a. 

When the factor $\lambda <\infty$, one can point out that  any vector $\tilde\fu=(\vec\fu,\nu)\in B_\lambda(\hat\fx)$ can be characterized by the following  inequality 
\begin{equation}
\label{eq:Charac_lambda}
\frac{\lambda}{2}\,\mu_2^2 + a_\lambda \left(\mu_1 - \frac{b_\lambda}{2}\right)^2 +\alpha\,|\nu|^{2}\leq \frac{c_\lambda}{4},
\end{equation}
where the scalar values $a_\lambda, b_\lambda, c_\lambda$ are all $1+\cO(1/\lambda)$. Hence the ball  $B_\lambda(\hat\fx)$ is an ellipsoid and is almost flat in the direction $\vec\fp_\theta^\perp$  due to the large value of  $\lambda /2$, as shown in Fig.~\ref{fig:Tissot}b for which we set $\alpha=1$. One can see that the unit ball $B_\lambda(\hat\fx)$ will converge to  $B_\infty(\hat\fx)$ in the sense of the Haussdorf distance, as $\lambda \to \infty$.

In Fig.~\ref{fig:UniformElasticas}b, we show the minimal paths derived from the metric $\kF_\lambda$. We denote by the red and black dots the source and end positions, respectively. In Fig.~\ref{fig:UniformElasticas}a, the  directions at each position are indicated by arrows. 

\subsection{Data-driven Finsler Elastica Metric}
\label{subsec:DataDrivenElastica}
In order to apply the Finsler elastica minimal paths for image segmentation and tubular structure centerline extraction, we should take into account the image data carried out by the orientation score $\psi:\bD\to\bR^+_0$ which is defined over the orientation-lifted domain $\bD$. 

The data-driven Finsler elastica metric $\cF_\lambda$ can be expressed by
\begin{equation}
\label{eq:DataFinslerElastica}	
\cF_\lambda(\hat\fx,\tilde\fu)=\exp\left(-\beta\frac{\psi(\hat\fx)}{\|\psi\|_\infty}\right)\kF_\lambda(\hat\fx,\tilde\fu),
\end{equation}
or
\begin{equation}
\label{eq:DataFinslerElastica2}	
\cF_\lambda(\hat\fx,\tilde\fu)=\left(1+\beta\frac{\psi(\hat\fx)}{\|\psi\|_\infty}\right)\kF_\lambda(\hat\fx,\tilde\fu),
\end{equation}
where $\beta$ is a positive constant. From Eqs.~\eqref{eq:DataFinslerElastica} and~\eqref{eq:DataFinslerElastica2},  we can see that the data-driven function $\Phi$ (see Eq.~\eqref{eq:DataDrivenBendingLength}) is replaced  by the orientation score-based functions $\exp(-\beta\psi(\hat\fx)/\|\psi\|_\infty)$ or $1+\beta\psi(\hat\fx)/\|\psi\|_\infty$. Note that in the experiments of this section, we make use of the metric defined in~\eqref{eq:DataFinslerElastica} for the data-driven Finsler elastica minimal path computation. 

The computation of the orientation score $\psi$ should be task-dependent.
We  respectively denote  by $\psi_{\rm edge}$ and $\psi_{\rm tube}$  the orientation score $\psi$ for image  segmentation and for tubular structure centerline extraction,  each of which can be estimated by steerable filters~\citep{freeman1991design}.

\begin{figure}[t]
\centering
\includegraphics[height=5cm]{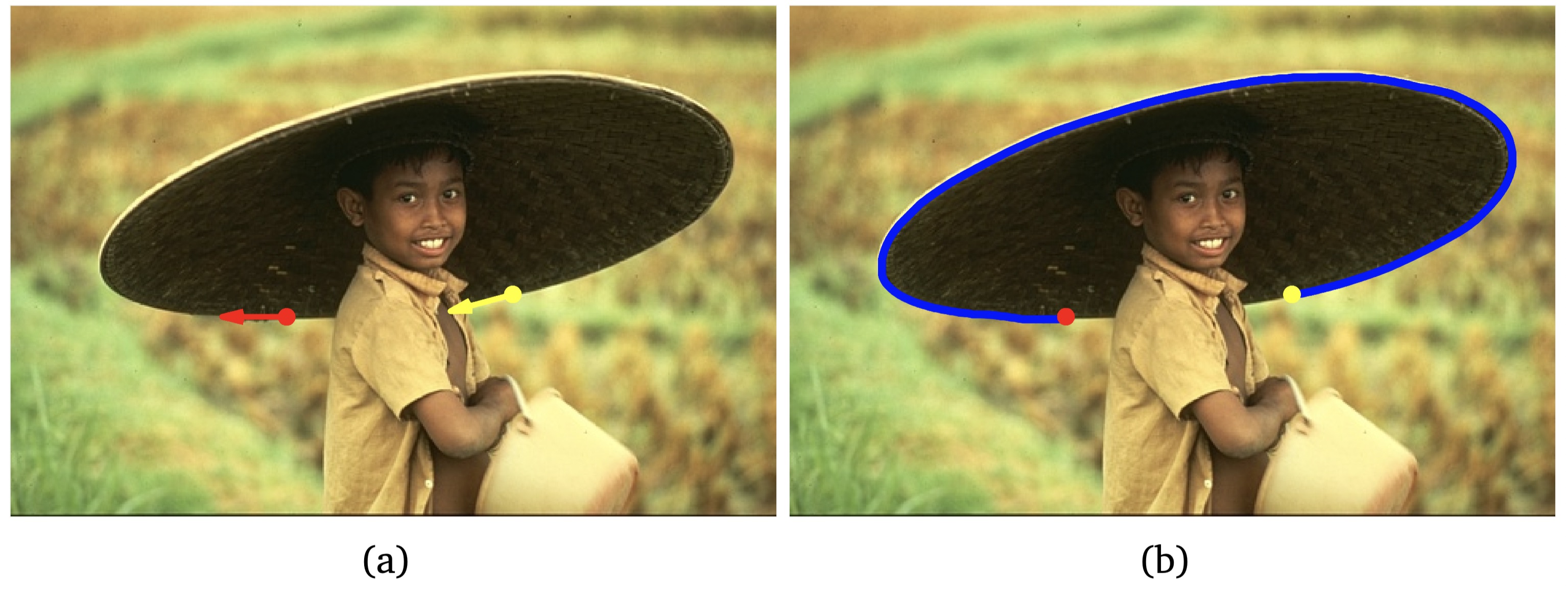}
\caption{An example for the minimal path derived from the data-driven Finsler elastica metric $\cF_\lambda$ based on the orientation score $\psi_{\rm edge}$. \textbf{a} The original image with source position (red dot) and end position (yellow dot). The arrows indicate the directions assigned to each position. \textbf{b} The obtained  minimal path which are indicated by the blue line. }
\label{fig:ElasticaEdge}
\end{figure}

\begin{figure}[t]
\centering
\includegraphics[height=6cm]{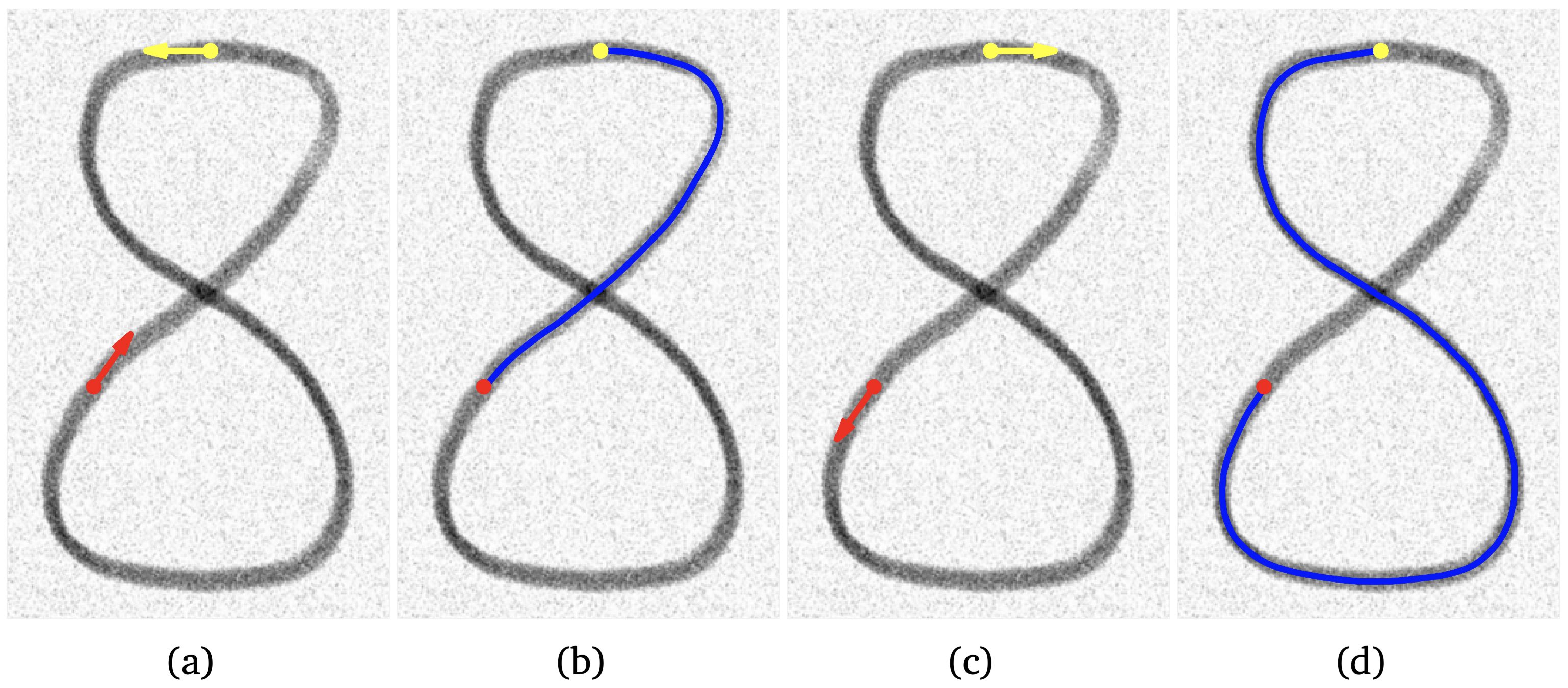}
\caption{Centerline extraction for a tubular structure based on data-driven Finsler elastica metric using the orientation score $\phi_{\rm tube}$. \textbf{a} and \textbf{c} Prescribed positions with arrows denoting the respective tangents. The red dots are the source positions and the yellow dots are the end positions. \textbf{b} and \textbf{d} The minimal paths corresponding to the initializations shown in \textbf{a} and \textbf{c}.}
\label{fig:ElasticaEight}
\end{figure}

\begin{figure}[t]
\centering
\includegraphics[height=10cm]{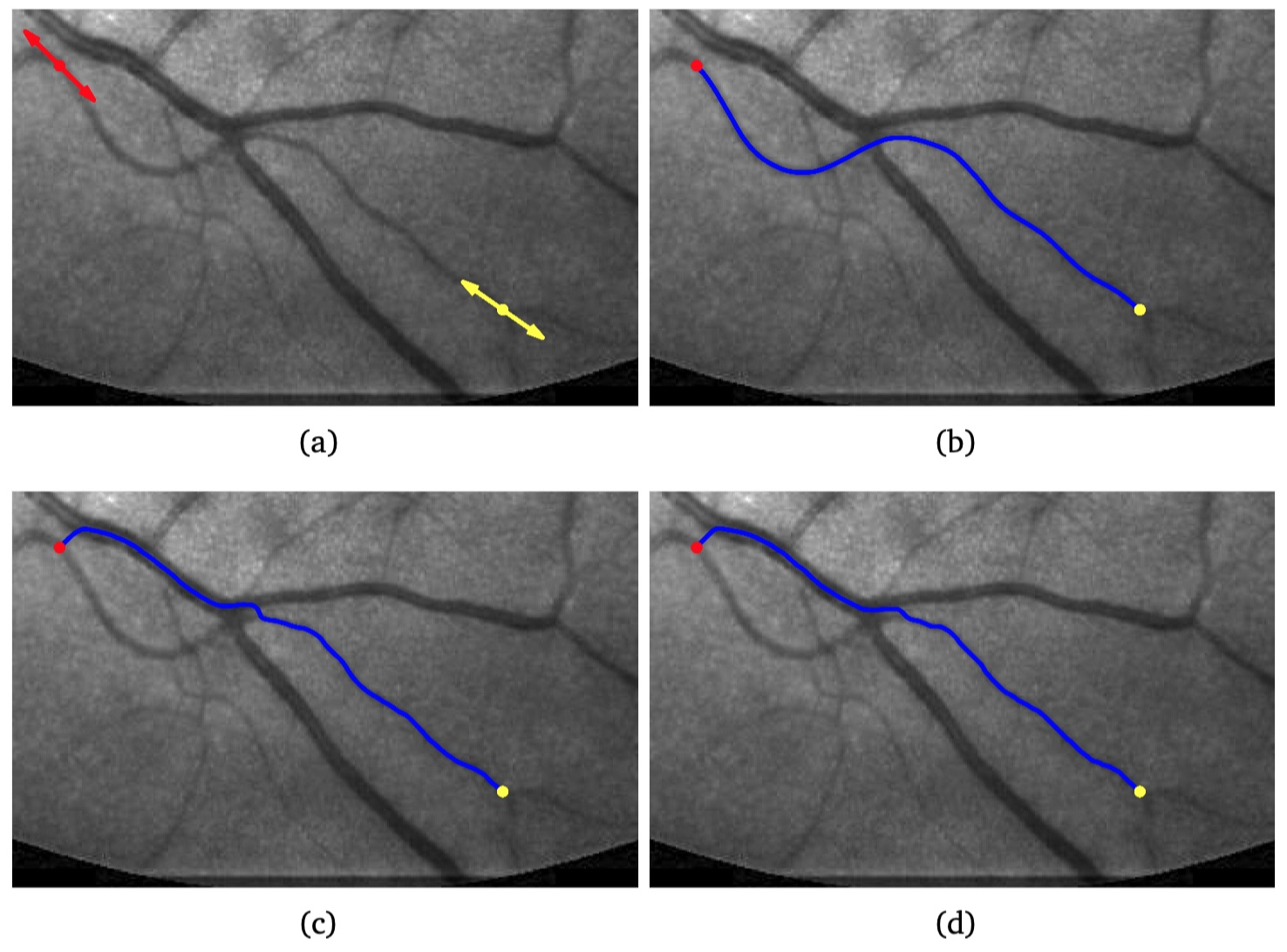}
\caption{Centerline tracking for a retinal vessel. \textbf{a} A retinal image patch. The red and yellow dots are the given source and end positions. The arrows denote the directions at the corresponding positions. \textbf{b} The extracted  minimal path associated to the data-driven Finsler elastica metric using the orientation score $\phi_{\rm tube}$. \textbf{c} and \textbf{d} The extracted  minimal paths associated to isotropic metric and anisotropic Riemannian metric, respectively.}
\label{fig:ElasticaVessel}
\end{figure}

Let $\mathbf I=(I_1,I_2,I_3):\Omega\to\bR^3$ be a RGB color image with three channels. For boundary detection and image segmentation applications, we use the $m$-order canny-like steerable filter~\citep{jacob2004design} with $m$ an odd integer to build the orientation score $\phi_{\rm edge}$. The $m$-order canny-like steerable kernel $\mathbf G^\theta_m$ can be expressed by
\begin{equation}
\mathbf G^\theta_m(\fx)=\sum_{a=1}^m\sum_{b=1}^a f_{a,b}(\theta)\frac{\partial^{a-b}}{\partial x^{a-b}} \frac{\partial^b}{\partial y^b}G_\sigma(\fx),
\end{equation}
where the function $G_\sigma$ is a Gaussian kernel with variance $\sigma$.   The coefficients $f_{a,b}$ are independent of the positions $\fx$ but relying on the orientation $\theta$ and we refer to~\citep{jacob2004design} for the computation of  $f_{a,b}$. Based on the kernels $\mathbf G^\theta_m$, the orientation score $\psi_{\rm edge}$ can be constructed  by 
\begin{equation}
\label{eq:OSEdge}
\psi_{\rm edge}(\fx,\theta)=\sum_{i=1}^3\left|\mathbf G^\theta_m\ast I_i(\fx) \right|. 
\end{equation}

The orientation score $\psi_{\rm tube}$ can be computed through a multi-scale tubular structure  enhancing filter such as the classical vesselness filter~\citep{frangi1998multiscale} or the optimally oriented flux (OOF) filter ~\citep{law2008three}. In this section, we use the OOF filter to derive the orientation score $\psi_{\rm tube}$, where the kernel $\mathbf{F}_r$ at a scale $r$ can be expressed as
\begin{equation}
\mathbf{F}_r(\fx)= \left(\mathbf 1_r\ast
\begin{pmatrix}
\partial_{xx}G_\sigma &	\partial_{xy}G_\sigma\\
\partial_{yx}G_\sigma & \partial_{yy}G_\sigma
\end{pmatrix}\right)(\fx).
\end{equation}
where $\mathbf 1_r$ is the indicator function  of a disk structure with radius $r$. In other words, the function $\mathbf 1_r$ can be defined as a step function relying on a radius $r$ such that
\begin{equation*}
\mathbf{1}_r(\fx)=
\begin{cases}
1,&\text{if~}\|\fx\|<r,\\
0,&\text{otherwise}.	
\end{cases}	
\end{equation*}
Then we can define a multi-scale response function at position $\fx$ and scale $r$ for a given image $I:\Omega\to\bR$ through the kernel $\mathbf{F}_r$
\begin{equation*}
\mathbf Q(\fx,r)=(\mathbf{F}_r\ast I)(\fx).	
\end{equation*}
Note that the response $\mathbf{Q}(\fx,r)$  is a symmetric matrix of size $2\times2$. Let us denote by  $\eta_1(\fx,r),\,\eta_2(\fx,r)$ the  two eigenvalues of the matrix $\mathbf{Q}(\fx,r)$ obeying that $\eta_2(\cdot)\geq\eta_1(\cdot)$. We suppose that the gray levels inside the tubular structure are lower than background. In this case,  for each point~$\fx$ we can estimate an optimal scale map $R_*:\Omega\to\bR^+$ by 
\begin{equation*}
R_*(\fx)=\underset{r}{\arg\max}\,\Big\{\eta_2(\fx,r)\Big\}.	
\end{equation*}
Then for a point $\fx$ inside the vessel region, the eigenvalues of the matrix $\mathbf{Q}(\fx,R_*(\fx))$ satisfy that  $\eta_2(\fx,R_*(\fx))\gg\eta_1(\fx, R_*(\fx))\approx0$. 
	
The orientation score $\psi_{\rm tube}$ can be computed  by   
\begin{equation}
\psi_{\rm tube}(\fx,\theta)=\max\{\langle\vec\fn_\theta,\mathbf  Q(\fx,R_*(\fx))\,\vec\fn_\theta\rangle,0\},
\end{equation}
where $\vec\fn_\theta=(-\sin\theta,\cos\theta)^T$ is a unit vector associated to an orientation $\theta\in\bS^1$.

The minimal geodesic paths derived from the data-driven Finsler elastica metric should be as smooth as possible and simultaneously follow the desired image features. In Fig.~\ref{fig:ElasticaEdge}, we illustrate an example for boundary detection using the data-driven Finsler elastica metric. In Fig.~\ref{fig:ElasticaEdge}a, the arrows indicate the directions at the given positions (denoted by red and yellow dots).  In Fig.~\ref{fig:ElasticaEdge}b, the obtained minimal geodesic path indicated by blue line can detect an image edge of high Euclidean length, which satisfies the properties  of the Finsler elastica minimal paths as discussed above.

In Fig.~\ref{fig:ElasticaEight}, we show the  geodesic paths derived from the data-driven Finsler elastica metric on a synthetic image. The arrows in Figs.~\ref{fig:ElasticaEight}a and~\ref{fig:ElasticaEight}c denote the corresponding directions assigned to the given positions indicated by the red and yellow dots. The geodesic paths shown in Figs.~\ref{fig:ElasticaEight}b and~\ref{fig:ElasticaEight}d are computed  using the same source  and end positions but opposite directions. One can see that these geodesic paths tend to  pass through the centerline of the target  tubular structure, and simultaneously try to avoid sharp turnings as much as possible. 

In Fig.~\ref{fig:ElasticaVessel}, we apply the Finsler elastica minimal path model for retinal vessel centerline tracking. In this experiment, only the positions are provided. Each position will be assigned two directions corresponding to the orientations maximizing  the orientation score $\psi_{\rm tube}$ over the space $[0,2\pi)$. In this case, the source position (resp. end position) will generate  two orientation-lifted source (resp. end) points. Hence we will obtain four pairs of orientation-lifted source and end  points, which correspond to four geodesic paths. Among them,  the geodesic  path with the  smallest geodesic distance value  will be chosen as the desired  path, see Fig.~\ref{fig:ElasticaVessel}b. For the purpose of comparison, we show the minimal paths respectively derived from the isotropic Riemannian metric~\citep{li2007vessels} and the  anisotropic Riemannian metric~\citep{benmansour2011tubular}, as shown in Figs.~\ref{fig:ElasticaVessel}c and Fig.~\ref{fig:ElasticaVessel}d. One can point out that these paths from the Riemannian metrics travel the segments belonging to different vessels, while the path from the Finsler elastica metric can track the correct vessel. Note that both of the Riemannian metrics used in this experiment are constructed by the OOF filter~\citep{law2008three}.

\section{Randers Minimal Paths for Region-based Active Contours}
\label{sec:EikonalMinimalPath}
In this section, we introduce a new minimal path model~\citep{chen2016finsler,chen2019eikonal} for solving the active contour problem involving a region-based homogeneity penalization term, using the minimal path and Eikonal PDE framework. 

\subsection{Hybrid Active Contour Model}
A hybrid active contour model~\citep{cohen1997avoiding,kimmel2003regularized,paragios2002geodesic,sagiv2006integrated} invokes an energy functional that is  made up of a region-based homogeneity penalization functional  $\rH_{\rm region}$ and  an edge-based weighted curve length $\rL_{\rm IR}$ (see Eq.~\eqref{eq:isotropicLength}), i.e.
\begin{equation}
\label{eq:HybridActiveContours}
E_{\rm hybrid}(\cC)=\alpha \rH_{\rm region}(\f1_{A_\cC})+\rL_{\rm IR}(\cC), 
\end{equation}
where $\alpha\in\bR^+$ is a parameter that controls the relative importance of the two terms $\rH_{\rm region}$ and $\rL$,  $\cC\in\Lip([0,1],\Omega)$ is a simple and closed curve and $\Omega$ represents the image domain. The function $\f1_{A_\cC}:\Omega\to\{0,1\}$ is the characteristic function of the open and bounded subset $A_\cC$ enclosed by $\cC$, which can be defined by
\begin{equation*}
\f1_{A_\cC}(\fx)=
\begin{cases}
1,&\forall\fx\in A_\cC,\\
0,&\forall\fx\in\Omega\backslash A_\cC.	
\end{cases}
\end{equation*}

The weighted curve length $\rL_{\rm IR}$ plays the role for regularization, which can be  measured using an edge-based potential $\cP:\Omega\to\bR^+$. In this section, we make  use of the following method to construct the potential
\begin{equation}
\label{eq:edgeTerm}
\cP(\fx)=\exp(\beta(\|g\|_\infty-g(\fx))),
\end{equation}
where $\beta\in\bR^+$ is a contrast parameter and the function $g$ represents the magnitude of the image gradient vector field, as defined in Eqs.~\eqref{eq:GrayLevelMagnitude} and~\eqref{eq:ColorMag}. One can see  that by Eq.~\eqref{eq:edgeTerm} the potential $\cP$ obeys that $\inf_{\fx\in\Omega}\{\cP(\fx)\}=1$ and appears to have low values around the image edges. 

Let  $\tilde\cC\in\Lip([0,1],\Omega)$  be a fixed  simple and closed curve that is parameterized in a clockwise order. Supposing that the region $A_\cC$ is close to $A_{\tilde\cC}$ and  by the differentiability of  the region-based functional $\rH_{\rm region}$, we can  obtain the following  approximation~\citep{chen2019eikonal} 
\begin{equation}
\label{eq:GateauxDerivative}
\rH_{\rm region}(\f1_{A_\cC})\approx c+\int_\Omega\rho_{\tilde\cC}(\fx)\f1_{A_\cC}(\fx)d\fx,
\end{equation}
where $\rho_{\tilde\cC}$ is the gradient of the functional $\rH_{\rm region}$ at $\f1_{A_{\tilde\cC}}$ and $c\in\bR$ is a scalar value that is independent to the characteristic function $\f1_{A_\cC}$. 

One can point out that only the second term in the right hand side of  Eq.~\eqref{eq:GateauxDerivative} relies on the characteristic function $\f1_{A_\cC}$ of the region $A_\cC$. For convenience, let us denote this  term by
\begin{align}
\cJ_{\tilde\cC}(\f1_{A_\cC})&=\int_\Omega\rho_{\tilde\cC}(\fx)\f1_{A_\cC}(\fx)d\fx\nonumber\\
\label{eq:ToMin}
&=\int_{A_\cC}\rho_{\tilde\cC}(\fx)\,d\fx.
\end{align}

In our model, solving the hybrid active contour problem through the curve evolution scheme amounts to iteratively searching for a family of successive simple and closed curves $\cC_{k}$ (indexed by $k$) such that the final curve $\cC_\infty$ can be used to depict the target boundary.  

In the $k$-th iteration ($k\geq 1$), the input is a curve $\cC_{k}$ obtained from the last iteration and the output  is a new simple and closed curve $\cC_{k+1}$. By setting $\tilde\cC:=\cC_{k}$, we can obtain the gradient $\rho_{\cC_{k}}$ of the functional $\rH_{\rm region}$ at $\f1_{\cC_{k}}$. Then the curve $\cC_{k+1}$ can be generated by solving 
\begin{equation}
\label{eq:mainProblem}
\min_{\cC\in\Xi_{\cC_k}}\{\alpha\cJ_{\cC_k}(\f1_{A_\cC})+\rL_{\rm IR}(\cC)\},
\end{equation}
where  $\Xi_{\cC_k}$ is a set of simple and closed curves. The construction of $\Xi_{\cC_k}$ relies on $\cC_k$ and will be introduced in the next section. 
In order to solve the problem~\eqref{eq:mainProblem} by the minimal path model and the Eikonal PDE, we transform the region-based functional $\cJ_{\cC_k}(\f1_{A_\cC})$ into a weighted curve length associated to a Randers metric using the divergence theorem, providing that the curve $\cC_{k+1}$ is chosen from the set $\Xi_{\cC_k}$.

\subsection{A Randers Metric Interpretation to the Hybrid Energy}
In the $k$-th iteration, let $U_{\cC_k}=\{\fx\in\Omega;d(\fx,\cC_k)<r\}$ be a tubular neighbourhood of the simple and closed curve $\cC_k$ with radius $r$, where $d(\fx,\cC)=\min_u\|\fx-\cC(u)\|$ denotes the Euclidean distance between $\fx$ and $\cC$. Then the set $\Xi_{\cC_k}$, which is made up of simple and closed curves, can be defined by
\begin{equation*}
\Xi_{\cC_k}=\{\cC\in\Lip([0,1],U_{\cC_k});A_{\cC_k}\backslash U_{\cC_k}\subset A_{\cC}\}.
\end{equation*}
For any  simple and closed curve $\cC\in\Xi_{\cC_k}$, one has $A_{\cC}\backslash U_{\cC_k}=A_{\cC_k}\backslash U_{\cC_k}$. This means that the region $A_\cC$ can be decomposed as 
\begin{align*}
A_\cC &= (A_{\cC}\backslash U_{\cC_k})\cup (A_{\cC}\cap U_{\cC_k})\\
&=(A_{\cC_k}\backslash U_{\cC_k})\cup (A_{\cC}\cap U_{\cC_k}).
\end{align*}
In this case, we can reformulate the region-based functional $\cJ_{\cC_k}$ as follows: 
\begin{align}
\cJ_{\cC_k}(\f1_{A_\cC})&=\int_{A_{\cC}\cap U_{\cC_k}}\rho_{\cC_k}(\fx)\f1_{U_{\cC_k}}(\fx)d\fx+\int_{A_{\cC_k}\backslash U_{\cC_k}}\rho_{\cC_k}(\fx)d\fx\nonumber\\
\label{eq:Div1}
&=\int_0^1\langle\vartheta_{\cC_k}(\cC(u)), \rN(u) \rangle \|\cC^\prime(u)\|du+\int_{A_{\cC_k}\backslash U_{\cC_k}}\rho_{\cC_k}(\fx)d\fx,
\end{align}
where $\f1_{U_{\cC_k}}$ is defined as the characteristic function of the tubular neighbourhood $U_{\cC_k}$, $\rN$ is the outward normal to the curve $\cC$, and  $\vartheta_{\cC_k}: \bR^2\to\bR^2$ is a vector field  which satisfies the following divergence equation over the domain $\bR^2$
\begin{equation}
\label{eq:DIVPDE}
\diver\vartheta_{\cC_k}=\rho_{\cC_k}\,\f1_{U_{\cC_k}}.
\end{equation}
One can always find such a vector field $\vartheta_{\cC_k}$ due to the existence of the solution to the  divergence equation~\eqref{eq:DIVPDE}.

In Eq.~\eqref{eq:Div1}, the second term  is independent to the curve $\cC$ and thus we only take into account the first term in the following calculation. Let $\mathbf{M}$ be the rotation matrix of angle $\pi/2$. Integrating the weighted curve length $\rL_{\rm IR}$ and the first term in Eq.~\eqref{eq:Div1}, we obtain  that
\begin{align}
&\alpha\int_0^1\langle\vartheta_{\cC_k}(\cC(u)), \rN(u) \rangle \|\cC^\prime(u)\|du+\int_0^1\cP(\cC(u))\|\cC^\prime(u)\|du\nonumber\\
=&\alpha\int_0^1\langle\mathbf{M}\vartheta_{\cC_k}(\cC(u)),\cC^\prime(u) \rangle du+\int_0^1\cP(\cC(u))\|\cC^\prime(u)\|du\nonumber\\
=&\alpha\int_0^1\kF_{\cC_k}(\cC(u),\cC^\prime(u))du,
\end{align}
where the metric $\kF_{\cC_k}$ has a form of
\begin{equation}
\label{eq:RegionRanders}
\kF_{\cC_k}(\fx,\vec\fu)=\cP(\fx)\|\vec\fu\|+\langle\alpha\,\mathbf{M}\vartheta_{\cC_k}(\fx),\vec\fu\rangle.	
\end{equation}
The positive definiteness condition~\eqref{eq:PositiveDefinitess} for $\kF_{\cC_k}$ requires that 
\begin{equation}
\label{eq:Positivity1}
\alpha\|\vartheta_{\cC_k}(\fx)\|<\cP(\fx),\quad \forall\fx\in U_{\cC_k}.
\end{equation}
However, the requirement~\eqref{eq:Positivity1} is difficult to satisfy. Hence we consider a new inequality formulated by
\begin{equation}
\label{eq:Positivity2}
\alpha\|\vartheta_{\cC_k}(\fx)\|<\inf_{\fy\in\Omega}\|\cP(\fy)\|=1.
\end{equation}

Now we have induced a Randers metric formulated in Eq.~\eqref{eq:RegionRanders}, which embeds the region-based homogeneity information as well as the image gradient features. This gives us the possibility of applying the minimal path framework to solve the region-based image segmentation problem. In next section, we will introduce a new vector field to approximate  $\vartheta_{\cC_k}$, such that the inequality~\eqref{eq:Positivity2} will always hold. 

\subsection{Practical Implementations}
In this section, we first introduce a method for computing the vector field $\vartheta_{\cC_k}$ used in the $k$-th iteration. In order to satisfy~\eqref{eq:Positivity2},  the vector field $\vartheta_{\cC_k}$ is expected to be as small as possible. Since the next evolving curve lies within the tubular neighbourhood $U_{\cC_k}$, we consider to find a vector field $\vartheta_{\cC_k}:U_{\cC_k}\to\bR^2$ by  solving the following minimization problem on the domain $U_{\cC_k}$
\begin{equation}
\label{eq:PDEDiv}
minimize~\int_{U_{\cC_k}} \|\vartheta_{\cC_k}(\fx)\|^2\,d\fx,\quad s.t.~\diver\vartheta_{\cC_k}=\rho_{\cC_k}~\text{on}~U_{\cC_k}.
\end{equation}
As discussed in~\citep{chen2016finsler,chen2019eikonal}, the value of $\sup_\fx\|\vartheta_{\cC_k}(\fx)\|$ is bounded by the area of the tubular neighbourhood $U_{\cC_k}$. Since the tubular neighbourhood $U_{\cC_k}$ acts as the search space for the evolving curves, reducing the area of $U_{\cC_k}$ may lead the proposed model being stuck in unexpected local minima. In order to address this issue,  we make use of  a new vector field $\varpi_{\cC_k}:U_{\cC_k}\to\bR^2$ derived by the nonlinear mapping $\varphi$ defined in Eq.~\eqref{eq:NonlinearApproximation} such that
\begin{equation}
\label{eq:ApproVector}
\varpi_{\cC_k}(\fx)=
\begin{cases}
\frac{\varphi(\tilde\alpha\|\vartheta_{\cC_k}(\fx)\|)}{\|\vartheta_{\cC_k}(\fx)\|}\vartheta_{\cC_k}(\fx), &\text{if~}\|\vartheta_{\cC_k}(\fx)\|\neq0,\\
\mathbf{0}, &\text{otherwise},
\end{cases}
\end{equation}
where $\tilde\alpha\in\bR^+$ is a constant. 

Replacing the vector field $\alpha\vartheta_{\cC_k}$ (see Eq.~\eqref{eq:RegionRanders}) by $\varpi_{\cC_k}$, one can obtain a new Randers metric:
\begin{equation}
\label{eq:ApproMetric}
\cF_{\cC_k}(\fx,\vec\fu)=\cP(\fx)\|\vec\fu\|+\langle\mathbf{M}\varpi_{\cC_k}(\fx),\vec\fu\rangle.
\end{equation}

In each  iteration $k$,  the weighted curve length measured along a curve $\cC\in\Xi_{\cC_k}$ can be formulated by 
\begin{equation}
\label{eq:variantLength}
\length_{\cF_{\cC_k}}(\cC)=\int_0^1\cF_{\cC_k}(\cC(u),\cC^\prime(u))du.
\end{equation}
For any point $\fx\in U_{\cC_k}$ the vector $\vartheta_{\cC_k}(\fx)$ is positively proportional to $\varpi_{\cC_k}(\fx)$ in the sense of the magnitude and of the direction. This gives the relevance between the original region-based active contour problem and the minimal path computation associated to the geodesic metric $\cF_{\cC_k}$. 
More  analysis for the reasonability of the use of the Randers metric $\cF_{\cC_k}$ can be found in~\citep{chen2019eikonal}. 

\begin{figure}[p]
\centering{
\includegraphics[width=14cm]{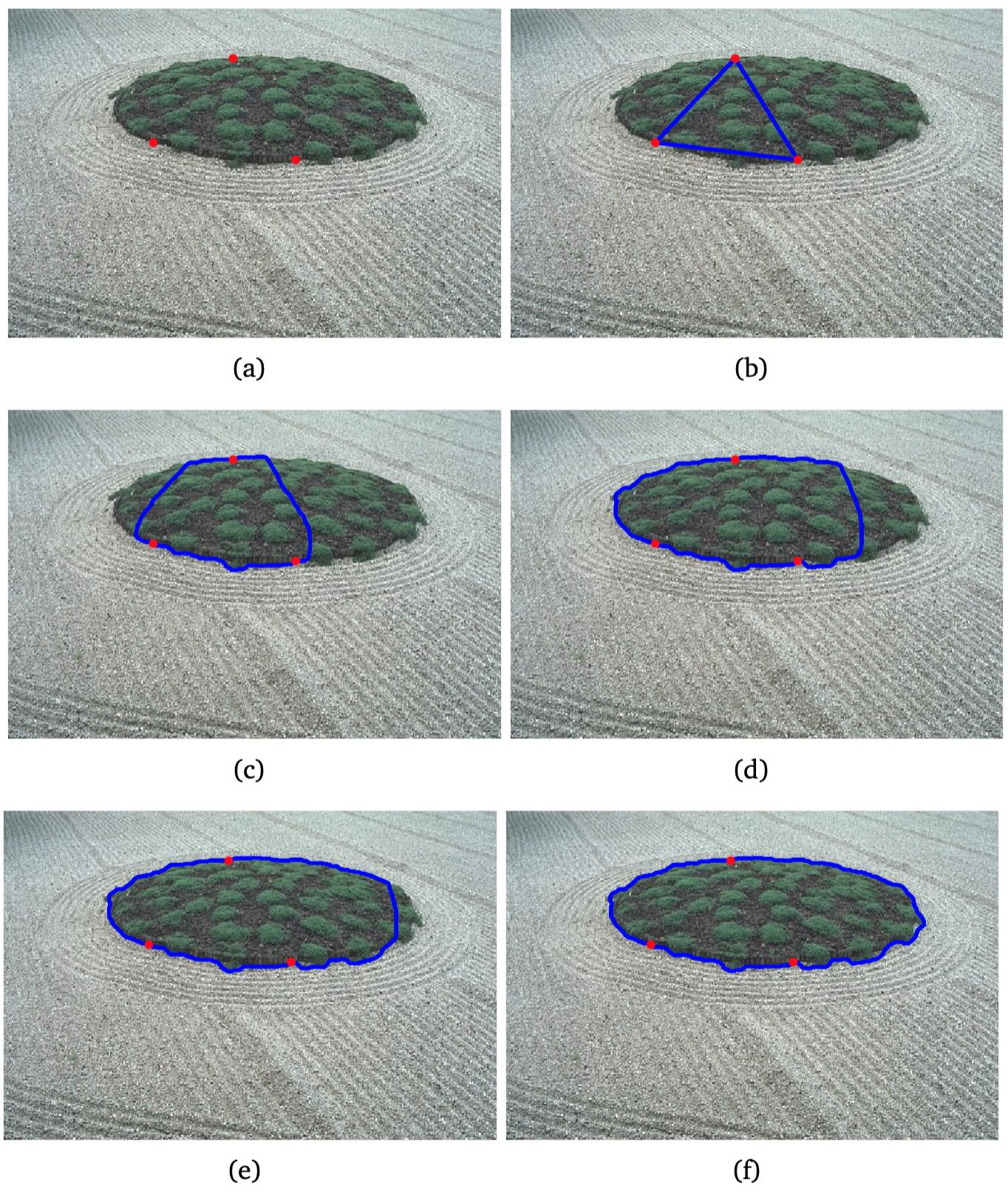}
}
\caption{Randers minimal paths for region-based image segmentation. \textbf{a} A set of prescribed vertices indicated by red dots. \textbf{b} The initial contour indicated by the blue line. \textbf{c} to  \textbf{e} Intermediate steps of the curve evolution processing.  \textbf{f} The final curve  generated by the concatenation of a set of Randers geodesic paths. One can see that all the closed curves indicated by blue lines pass through the red dots.}
\label{fig:RegionRanders1}	
\end{figure}

\subsection{Application to Image Segmentation}
We make use of the piecewise constant homogeneity term~\citep{chan2001active} to build the Randers metric $\cF_{\cC_k}$ in Eq.~\eqref{eq:ApproMetric} and  illustrate how to  apply $\cF_{\cC_k}$ for image segmentation. The objective is to search for a family of successive  $\cC_k$, each of which is simple and closed, to depict the object boundary, as $k\to\infty$. 

The region-based homogeneity term in the piecewise constant variant of the Mumford-Shah model can be expressed  as follows
\begin{equation*}
\rH_{\rm region}(\f1_{A_\cC},\mu_{\rm in},\mu_{\rm out})=\int_\Omega(I-\mu_{\rm in})^2\f1_{A_\cC} d\fx+\int_\Omega(I-\mu_{\rm out})^2(1-\f1_{A_\cC})d\fx.
\end{equation*}
Recall that in the $k$-th iteration of the curve evolution scheme, the input is the curve  $\cC_{k}$ ($k\geq1$) and the output is $\cC_{k+1}$. Note that when $k=1$,  $\cC_1$ is the initial curve.  We can estimate the gradient $\rho_{\cC_k}$ of the region-based functional $\rH_{\rm region}$, which can be formulated by
\begin{equation}
\label{eq:CVGradient}
\rho_{\cC_k}=\left(I(\fx)-\mu_{\rm in}[\cC_{k}]\right)^2-\left(I-\mu_{\rm out}[\cC_{k}]\right)^2,
\end{equation}
where $\mu_{\rm in}[\cC_{k}]$ and $\mu_{\rm out}[\cC_{k}]$ are respectively the mean gray levels inside and outside the curve $\cC_{k}$.
 
Based on the gradient $\rho_{\cC_k}$ and the tubular neighbourhood $U_{\cC_k}$, one can solve the PDE-constrained  minimization problem~\eqref{eq:PDEDiv} to obtain the vector field $\vartheta_{\cC_k}$ and also to obtain $\varpi_{\cC_k}$. Finally, the Randers metric $\cF_{\cC_k}$ can be constructed using Eq.~\eqref{eq:ApproMetric}. 

\emph{Interactive image segmentation method}.
By solving the gradient descent ODE, one can obtain an open geodesic path between two points within a given domain. However, for image segmentation, the objective is to seek simple and closed curves. Thus we need to solve an issue: how to derive  a closed and simple curve $\cC_{k+1}$ in each iteration $k$, providing that the Randers metric $\cF_{\cC_k}$ is given.
In this section, we consider an interactive segmentation  scheme to obtain the evolving curve $\cC_{k+1}$, which depends on a set $\{\fq_i\}_{1\leq i\leq m}$ of $m$ ($m\geq3$)  prescribed vertices distributed on the target boundary in a clockwise order. These vertices will  be fixed in the course of curve evolution. Each evolving curve $\cC_k$ is supposed to pass through each of these  vertices in a clockwise order. The initial curve $\mathcal{C}_1$ is  a polygon which is generated using the vertices  $\mathbf{q}_i$. The simple and closed curve $\cC_{k+1}$ can be generated by the end-to-end concatenation of a set of geodesic paths obtained using $\cF_{\cC_k}$. Each of these geodesic paths, denoted by $\cG_{i,k+1}$ with $1\leq i\leq m$, connects a vertex $\fq_i$ to its successive one.

Now we present the method for computing the geodesic paths $\cG_{i,k+1}$ to obtain the curve $\cC_{k+1}$. Recall that the input curve $\cC_k$ passes through all of the fixed vertices $\{\fq_i\}_{1\leq i\leq m}$. In this case, we are able to decompose $\cC_k$ into a set $\{\cG_{i,k}\}_{i}$ of $m$ subpaths. Each path $\cG_{i,k}$, defined over the integral $[0,1]$, links a vertex $\fq_i$ to its successive one $\fq_{i+1}$ for $i<m$ or to $\fq_1$ for $i=m$. Following that  the tubular neighbourhood $U_{\cC_k}$ can be decomposed into a family  of disjoint subdomains $\mathscr{U}_i\subset U_{\cC_k}$, each of which can be regarded as a narrowband region (with radius $r$) involving the open curve $\cG_{i,k}$. In essence, these regions $\mathscr{U}_i$ ($i=1,2,\cdots m$) can be identified by using the Voronoi index map over the domain $U_{\cC_k}$ associated to the open curves $\cG_{i,k}$ ($1\leq i\leq m$) as introduced in~\citep{chen2019eikonal}. In other words,  a point $\fx\in\mathscr{U}_i$ implies that $d(\fx,\cG_{i,k})<d(\fx,\cG_{j,k}),\,\forall j\neq i$.

With these definitions in hands, we can track a  geodesic path $\cG_{i,k+1}$ in the corresponding  region $\mathscr{U}_i\cup\{\fq_i,\fq_{i+1}\}$ if $i<m$, or $\mathscr{U}_m\cup\{\fq_m,\fq_{1}\}$ if $i=m$. The use of these regions  $\mathscr{U}_i\cup\{\fq_i,\fq_{i+1}\}$ for geodesic path computation is to avoid the curve self-crossing issue and also to reduce the computation time. The decomposition of the tubular neighbourhood $U_{\cC_k}$ can be efficiently implemented by the Fast Marching method with a Voronoi index map estimation procedure.  Once all of the geodesic paths $\cG_{i,k+1}\,(1\leq i\leq m)$ are generated, we can construct a simple and  closed curve $\cC_{k+1}$ by the concatenation  of the paths $\cG_{i,k+1}$ in an end-to-end manner. Finally, the curve $\cC_{k+1}$ is regarded as the output of the $k$-th iteration. 

Note that  for a pair of successive  vertices $(\fq_i,\fq_{i+1})$, the extracted geodesic path $\cG_{i,k+1}$ is the globally optimal curve for the length $\length_{\cF_{\cC_k}}$ (see Eq.~\eqref{eq:variantLength}) within the domain $\mathscr{U}_i\cup\{\fq_i,\fq_{i+1}\}$.

In Fig.~\ref{fig:RegionRanders1}, we show an example for this interactive image segmentation procedure providing that a set of $3$ ordered vertices are given. These vertices are illustrated by red dots, see Fig.~\ref{fig:RegionRanders1}a.  In Fig.~\ref{fig:RegionRanders1}b, the blue straight segment lines connecting each pair of red dots represent the initial curve. In Figs.~\ref{fig:RegionRanders1}c to~\ref{fig:RegionRanders1}e, we show the intermediate curve evolution results, where the blue contour is generated by the concatenation of a set of the Randers minimal paths.  The final curve is shown in Fig.~\ref{fig:RegionRanders1}f. Again, the vertices denoted by red dots are fixed  during the curve evolution. 

\begin{remark}
The method presented in this section assumes that the vertices are provided by user and are fixed during the curve evolution. As a matter of fact,  in each iteration $k$, these vertices can be sampled from the input curve $\cC_k$. In this case, one can initialize the proposed model by providing a closed curve and the vertices will evolve  in the course of the curve evolution processing~\citep{chen2016finsler}.
\end{remark}

\begin{remark}
Note that a variant of the edge-based balloon active contour model can also be addressed by the framework proposed in this section. The balloon force $\mathbf{F}_{\rm balloon}$ in Eq.~\eqref{eq:BalloonForce} can be obtained by minimizing the following term
\begin{equation}
\label{eq:VariantBalloon}
\int_{A_\cC}-1\,d\fx,
\end{equation}
yielding a curve evolution flow 
\begin{equation}
\label{eq:VariantBalloonFlow}
\frac{\partial\cC_\tau}{\partial\tau}=\rN_\tau.
\end{equation}
Recall that $\rN_\tau$ is the outward normal to $\cC_\tau$.

Now we can establish  a variant form of the isotropic geodesic active contour functional~\citep{caselles1997geodesic,yezzi1997geometric} with the balloon force term
\begin{equation}
\label{eq:balloon}
E_{\rm balloon}	(\cC)=\alpha\,\int_{A_\cC}-1\,d\fx+\rL_{\rm IR}(\cC).
\end{equation}
In this case, for a given simple and closed curve $\tilde\cC$, the gradient of $E_{\rm balloon}$ thus turns out to be a constant function, i.e. $\rho_{\tilde\cC}\equiv1$. This edge-based active contour model thus can be solved by the Randers metric-based model proposed in this section. 

Furthermore, an alternative choice for the regularization term in Eqs.~\eqref{eq:HybridActiveContours} and~\eqref{eq:balloon} can be constructed using an anisotropic Riemannian metric $\kR$ (see Eq.~\eqref{eq:RiemannianMetric}), which allows to take into account the path directions for image segmentation~\citep{chen2017anisotropic,chen2019eikonal}.
\end{remark}

\section{Conclusion}
In this chapter, we review the methods for the construction  of the Riemannian and Randers metrics in terms of various active contour terms. We show that by these  metrics  the edge- and region-based active contour problems  and  the Euler-Mumford elastica problem can be efficiently solved by the minimal path framework based on the Eikonal partial differential equation.  Moreover, we also exploit the relevant applications of minimal paths associated to the proposed metrics in image analysis such as image segmentation, boundary detection as well as  tubular structure centerline  tracking, which are able to blend the benefits from the minimal path framework and the active contour models.  The well-established fast marching methods, which are the Eikonal solvers, allow efficient and practical implementations for these applications. 

\section*{Acknowledgement}
The authors thank Dr. Jean-Marie Mirebeau from   Universit\'e Paris-Saclay for his fruitful discussion and regular collaboration. This research has been partially funded by Roche pharma (project AMD\_short) and by a grant from the French Agence Nationale de la Recherche ANR-16-RHUS-0004 (RHU TRT\_cSVD).	Figures~\ref{fig:Tissot} are reprinted by permission from Springer Nature Customer Service Centre GmbH:  Springer Nature, International Journal of Computer Vision, Global minimum for a Finsler elastica approach, \emph{Da Chen, Jean-Marie Mirebeau and Laurent D. Cohen}~\citep{chen2017global}.

\bibliographystyle{apalike}  
\bibliography{MinimalPathsApplications}

\end{document}